\documentclass[journal]{IEEEtran}
\usepackage{graphicx}%figure package
\usepackage{float}
\usepackage{subfigure}

\usepackage{amssymb}
\usepackage{array}
\usepackage{amsmath}
\usepackage{url}
\usepackage{multirow}
\usepackage{booktabs}
\usepackage{threeparttable}
\usepackage{fixltx2e}
\usepackage{float}
\usepackage{booktabs}
\usepackage{graphicx}
\usepackage{diagbox}
\usepackage{mathtools}
\usepackage{amsmath}
\usepackage{bbding}
\usepackage{pifont}
\usepackage{wasysym}
\usepackage{tikz}
\usepackage{calc}
\usepackage{amssymb}
\usepackage{multirow}
\usepackage{rotating}
\usepackage{colortbl}
\usepackage{color}
\usepackage{ragged2e}
\usepackage[colorlinks,linkcolor=blue,anchorcolor=blue,citecolor=blue]{hyperref}
\usepackage{cite}
\usepackage{gensymb}
\usepackage{makecell}
\usepackage{comment}

\begin{document}
\title{Soft Masked Transformer for Point Cloud Processing with Skip Attention-Based Upsampling}
\author{
Yong~He,
        Hongshan~Yu,
        Chaoxu Mu,\\
        Mingtao Feng,
        Tongjia Chen,
        Zechuan Li,
        Anwaar Ulhaq,
        Ajmal~Mian. % <-this % stops a space
\thanks{}
\thanks{Yong He, Chaoxu Mu are with the Artificial Intelligence School, Anhui University, 11 Jiulong Road, Shushan Dist., 230601, Hefei, China.}
\thanks{Hongshan Yu,  Tongjia Chen are with the National Engineering Laboratory for Robot Visual Perception and Control Technology, College of Electrical and Information Engineering, Hunan University, Lushan South Rd., Yuelu Dist., 410082, Changsha, China. This work was partially supported by the National Natural Science Foundation of China (Grants U2013203, 61973106).}
\thanks{Mingtao Feng is with the Artificial Intelligence School, Xidian University, 266 Xinglong Section, Xifeng Road, 710071, Xian, China. }
\thanks{Ajmal~Mian is with the Department of Computer Science, The University of Western Australia, WA 6009, Australia. Ajmal Mian is the recipient of an Australian Research Council Future Fellowship Award (project number FT210100268) funded by the Australian Government.}
\thanks{Anwaar Ulhaq is with Central Queensland University, Sydney Campus, Australia.}
}

% \markboth{IEEE Transactions on MultiMedia, Class Files,~Vol.~14, No.~8, August~2021}%
% {Shell \MakeLowercase{\textit{et al.}}: Bare Demo of IEEEtran.cls for IEEE Journals}

\markboth{IEEE Transactions on Automation Science and Engineering}%
{Shell \MakeLowercase{\textit{et al.}}: Bare Demo of IEEEtran.cls for IEEE Journals}

\maketitle
\begin{abstract}
Point cloud processing methods leverage local and global point features %at the feature level 
to cater to downstream tasks, yet they often overlook the task-level context inherent in point clouds during the encoding stage. We argue that integrating task-level information into the encoding stage significantly enhances performance. To that end, we propose SMTransformer which incorporates task-level information into a vector-based transformer by utilizing a soft mask generated from task-level queries and keys to %re-weight 
learn the attention weights. Additionally, to facilitate effective communication between features from the encoding and decoding layers in high-level tasks such as segmentation, we introduce a skip-attention-based up-sampling block. This block dynamically fuses features from various resolution points across the encoding and decoding layers. 
To mitigate the increase in network parameters and training time resulting from the complexity of the aforementioned blocks, we propose a novel shared point position encoding strategy. This strategy allows various transformer blocks to share the same position information over the same resolution points, thereby reducing network parameters and training time without compromising accuracy.
Experimental comparisons with existing methods on multiple datasets demonstrate the efficacy of SMTransformer and skip-attention-based up-sampling for semantic segmentation task. In particular, we achieve state-of-the-art semantic segmentation results of \textcolor{black}{73.9\% mIoU on S3DIS Area 5} \textcolor{black}{and 62.4\% mIoU on SWAN dataset}. %Our code is available at
% https://github.com/hnuhyuwa/SMTransformer.

% Anwaar: I have modified Abstract sentences and fixed a few typos 

\end{abstract}

\vspace{-1mm}
\begin{IEEEkeywords}
Deep learning, 3D point clouds, Soft mask, Transformer, Attention-based up-sampling.

\vspace{1mm}
\textcolor{black}{\textbf{\textit{Note to Practitioners—}Point cloud processing underpins automation tasks such as robotic perception, navigation, and inspection, where accurate 3D understanding is essential. Existing methods often prioritize vision benchmarks while overlooking automation needs like efficiency on limited hardware and robustness in real-world environments. The proposed SMTransformer embeds task-level guidance into feature learning and employs skip-attention up-sampling to improve segmentation accuracy with practical efficiency. It is well-suited for robotic manipulation, autonomous driving, and inspection applications. Current limitations include reliance on GPUs and sensitivity to extreme density variations. Future work will target edge-device deployment and multi-task extensions.}}

% \textbf{\textit{Note to Practitioners—}Point cloud processing is critical for automation tasks like robotic navigation and perception of autonomous vehicles, where accurate interpretation of 3D environments is essential. However, existing methods often struggle to balance computational efficiency with performance when handling complex, real-world scenes. This limitation arises because they neglect task-specific context during feature extraction and use rigid up-sampling strategies that limit adaptability. Our work addresses these challenges by introducing a novel framework that dynamically aligns feature learning with downstream objectives while reducing computational overhead.}

% \textcolor{black}{\textbf{}}

% \textbf{These innovations enable higher-quality semantic segmentation  in indoor/outdoor scans while maintaining practical training times. Applications include robotic manipulation in warehouses, infrastructure inspection, and autonomous driving systems requiring real-time 3D scene understanding. Current limitations include reliance on GPU acceleration for optimal inference speed and sensitivity to extreme point density variations. Future extensions could integrate our approach with edge-device optimization techniques or extend it to multi-task frameworks (e.g., joint detection and segmentation). By bridging task awareness and efficiency, this work offers a versatile tool for engineers designing intelligent systems based on robust 3D perception.}
\end{IEEEkeywords}

\section{Introduction}
With the rapid development of 3D sensors such as LiDARs and depth cameras, the accessibility of 3D point cloud data has dramatically increased. 
% Consequently, its importance is growing across various applications, 
\textcolor{black}{Consequently, its role in automation-oriented applications has expanded to areas such as autonomous driving \cite{10620438}, robotics\cite{wang2024mnat }, and industrial automation\cite{ren2022uni3da}, where reliable 3D perception is critical for decision-making and control.} Techniques for processing 3D point clouds have attracted the interest of many researchers. Effectively learning features from 3D point clouds is challenging due to its irregular nature, as opposed to regular grid like structure of images. Inspired by the success of grid convolutions, several methods have been proposed to transform irregular point clouds into regular representations, including projected images, or voxels. However, the discretization process inevitably sacrifices significant geometric information.

Early approaches learn features from raw point clouds by employing a shared multi-layer perceptron (MLP) on each point and use symmetric functions (e.g., max-pooling) to aggregate the most prominent features over the receptive field. However, this design ignores local structures crucial for shape representation. Inspired by 2D grid convolutions, point convolutions employ correlation functions to quantify relationships among points, enabling the model to leverage \textcolor{black}{both local features (e.g., edges, corners, surfaces) for robotic perception and contextual cues (e.g., scene layout) essential in automation tasks.} Some point convolutions use weight functions to learn weights from various local point geometric connections, such as point coordinates, coordinate differences, distances, etc., \cite{thomas2019kpconv}. Others associate coefficients (derived from point coordinates) \cite{wu2019pointconv} with weight functions to adjust the learned weights.

In contrast to point convolution, the point transformer focuses on learning feature connections and attention maps between point features through a scalar or vector-based attention mechanism globally \cite{feng2020point}, \cite{yan2020pointasnl} or over a local receptive field \cite{wu2022point}. Recently, with the help of position encoding, the aggregation ability of point transformers has seen significantly improvement. While point transformers are powerful learners, they generally fall short of simultaneously exploiting local features and global context, failing to establish communication between the two. Current point convolution and transformer based approaches strive to enhance the encoder and decoder designs to facilitate proficient point connection learning for improved performance on  downstream tasks such as segmentation and classification. However, these methods often overlook capturing the task-level information of the entire point cloud. \textcolor{black}{For instance, task-level information in the semantic segmentation task refers to the semantic priors (e.g. class scores) provided by the complete set of class labels, which are crucial for disambiguating locally similar but semantically different structures}

To address the aforementioned issues, we propose a Soft Masked Transformer (\textbf{SMTransformer}) that incorporates task-level contextual information into attention mechanisms and utilizes this information to guide local feature learning softly. Specifically, the SMTransformer predicts \textcolor{black}{class score} keys and queries over the global receptive field and then obtains a soft mask from these keys and queries to re-weight the local attention map.

To establish communication between encoder and decoder layers, conventional point transformer networks typically incorporate skip connections and fusion within the up-sampling block of the decoder layer. However, this straightforward fusion approach may not dynamically blend features from the encoder and decoder layers. Furthermore, these operations are limited to the same resolution point cloud i.e. there is no communication between points at different resolutions. To address this challenge, we introduce a novel Skip-Attention-based up-sampling Block (\textbf{SAUB}). This block initially \textcolor{black}{augments} the resolution of low-resolution points \textcolor{black}{at the position level}, while preserving their original features. Subsequently, it learns attention maps to establish meaningful connections \textcolor{black}{by skip attention} between low-resolution and high-resolution point features.

Moreover, in conventional networks, different point transformer blocks often employ distinct position encoding information for the same resolution. This increases the network parameters and the training time. Intuitively, the same resolution points in the network have the same position information, a characteristic independent of the number and location of point transformers. To summarize, our contributions are threefold:

\begin{itemize}
\vspace{1mm}

\item We propose a Soft Masked Transformer block, which integrates task-level information into the attention mechanism, enhancing its effectiveness for  semantic segmentation.

\item We introduce a Skip Attention-based Up-sampling block, which dynamically combines features from different resolution points across the encoding layers, improving the model's ability to capture contextual information.

\vspace{1mm}

\item We present a shared point position encoding strategy, which reduces network parameters by 24\% and training time by 33\%. This strategy enhances the efficiency of the network without sacrificing performance.

\end{itemize}
\indent 

We conduct extensive experiments on benchmark datasets to showcase the effectiveness of our proposed method and their robust generalization across various tasks, including indoor semantic segmentation, outdoor semantic segmentation
% , and object classification. 
Our method consistently achieves competitive results compared to existing point transformer-based approaches. Particularly noteworthy is our method's achievement of state-of-the-art performance, attaining the remarkable mIoU of \textcolor{black}{73.9\%}  on the S3DIS Area 5 and 62.4\% on the SWAN dataset without \textcolor{black}{\textit{using any extra data}.}

% Anwaar: Some typos are fixed. 

\vspace{-2mm}
\section{Related Work}\label{section2}

\subsection{Transformer-based Methods} \textcolor{black}{Aiming to maximize the preservation of geometric information in point clouds, state-of-the-art methods prefer to directly process the raw point clouds. The development of the most important unit (i.e., local aggregation) in the point cloud processing network can be broadly divided into three categories, including MLP-, Convolution- and Transformer-based approaches. MLP-based methods \cite{qi2017pointnet++, ma2021rethinking} employ shared MLPs to extract point-wise features and typically rely on symmetric functions (e.g., max-pooling) to aggregate them. While simple and efficient, they struggle to capture rich local structures that are critical for vision tasks. Motivated by 2D convolution, point convolution methods \cite{wu2019pointconv, Lei_2020_CVPR} extend convolutional operations to irregular point sets or point graphs. These approaches dynamically generate convolutional kernels based on local geometric relationships. However, since the weights are learned primarily from low-level geometric cues (e.g., coordinates or relative positions), their ability to capture higher-level contextual dependencies remains limited.}

Unlike convolution-based methods, which learn convolutional weights from low-level point coordinates, attention mechanisms learn attention weights from the connections between point features, thereby exploiting high-level contextual information. Motivated by the success of attention mechanisms in natural language processing and image processing tasks, early methods applied self-attention to global points through scalar dot-product \cite{feng2020point, yan2020pointasnl}, but suffered from high computational costs. These early attention-based methods did not demonstrate superior performance due to the lack of employing position encoding.

The Point Transformer \cite{zhao2021point} introduces local vector attention to the local points. Additionally, it emphasizes the significance of point position encoding. Subsequent work, Point Transformer V2 \cite{wu2022point}, incorporates multi-grouping into vector attention inspired by the multi-head strategy. It also enhances position encoding by introducing an additional multiplier to the relation vector, which facilitates learning complex positional relations. To exploit long-range contextual information, the Stratified Transformer \cite{lai2022stratified} densely selects nearby points over a cubic window and sparsely selects distant points. This stratified strategy enlarges the effective receptive field without incurring too much computational overhead. However, by using a window-based approach, the Stratified Transformer focuses on an expanded local region rather than the global region. Moreover, hyper-parameters such as the window size and number of distant points must be optimized for different datasets with varying densities.

\vspace{-2mm}
\subsection{Up-sampling in 3D Point Clouds}
The hierarchical architecture of a network is instrumental in learning long-range contextual information through down-sampling and grouping operations. Up-sampling involves interpolating new points between known points and adjusting the features of these new points based on their mutual distance to propagate the learned context features to each point. Compared to the extensive research on down-sampling \cite{ hu2020randla, yan2020pointasnl} and grouping operations \cite{ feng2020point}, few works specifically emphasize the up-sampling operation. In the Point Transformer \cite{zhao2021point}, the transition up module uses an interpolation operation to recover new point features from known point features through indexing and then integrates these features with those from the encoding layer via a skip connection. Similarly, in Point Transformer V2 \cite{wu2022point}, the fusion of encoding and decoding layer features is achieved through a skip connection. The distinguishing factor here is that the new point features are unpooled by grid unpooling instead of interpolation. While these methods are simple, they lack semantic awareness and ignore the contextual connection between the encoding and decoding layers.

\begin{comment}
    I have corrected grammatical mistakes and improved the language. However, I am less convinced by the literature survey. My point is to mention the literature that highlights the research gaps that motivate the work in this paper. I see a missing link between the literature and the motivation of this work. Another observation: the top three subsections, A, B, and C, are mentioned but no overview is provided explaining why these three areas are related to this paper. I suggest adding a few sentences after the heading "Related Work" to provide context for these subsections.
\end{comment}

Our survey highlights several gaps of in current point cloud processing methods. Their fundamental unit, such as point MLP, convolution, transformer, completely ignores task-level information which then leads to sub-optimal performance in downstream tasks. Moreover, the upsampling methods do not effectively communicate between the encoding layers over different resolution points to refine the context information for high-level tasks. Motivated by these gaps, we propose a soft-masked transformer and skip-attention-based upsampling. Moreover, we propose shared point position encoding to reduce the network parameters and training time.

% \vspace{-2mm}
\section{Method}
\label{section3}
We revisit the two classical local vector attention-based point transformers, namely Point Transformer and Point Transformer V2, in Section \ref{section3_subsection1}. Then, we present the SMTransformer block in Section \ref{section3_subsection2}, followed by our skip attention-based up-sampling block in Section \ref{section3_subsection3}. We introduce the shared point position encoding strategy in Section \ref{section3_subsection4}, and finally, in Section \ref{section3_subsection5}, we provide details about our network.

% \vspace{-3mm}
\subsection{Rethinking Vector Attention based Transformer}\label{section3_subsection1}

Denote a point cloud $p_{i}\in\mathbb{R}^{{N\times3}}$ (where $p_i$ defines point positions) and its corresponding features $f_{i}\in\mathbb{R}^{{N\times C}}$. $f_i$ is a feature vector that may contain attributes such as normal vectors and colour of the surface. $N$ and $C$ are the number of points and feature channels, respectively. We denote the K neighbors of $p_{i}$ as $p_{ij}\in\mathbb{R}^{{N\times K\times3}}$ and their corresponding features as $f_{ij}\in\mathbb{R}^{{N \times K\times C}}$. The position over the local receptive field can be expressed as $\Delta p_{ij}$. Point Transformer\cite{zhao2021point} on point cloud $p_i$ can be expressed as,
\vspace{-2mm}
\begin{equation}\label{eq:pointtransformer}
\mathcal{G}_{i}=\sum_{j=1}^{K}\mathcal{A}\big((K_{ij}^f \ominus Q_i^f) \oplus \delta_b \big) \odot \big(V_{ij}^f \oplus \delta_b \big), 
\vspace{-2mm}
\end{equation}
\noindent where $Q_i^f\in\mathbb{R}^{{N\times C}}$ is the query matrices of $f_i$. $K_{ij}^f$ and $V_{ij}^f\in\mathbb{R}^{{N\times K\times C}}$ are the key and value matrices of $f_{ij}$. The subtraction operation $\ominus$ between $Q_i^f$ and $K_{ij}^f$ is performed via broadcasting to ensure dimension compatibility. $\delta_b = \delta_b(\Delta p_{ij})$ is the position encoding bias. $\mathcal{A}(\cdot)$ denotes the local vector attention function, implemented by MLP, followed by the softmax. $\sum$ stands for the symmetric operation (SOP)  (e.g. summation) and $\odot$ is the element-wise multiplication operation. $\oplus$ is the element-wise addition operation and $\mathcal{G}_{i}$ is the output feature of transformer. 

To exploit the more complex geometric relationships between points, Point Transformer V2\cite{wu2022point} strengthens the position encoding with an additional multiplier  to the relation feature vector (i.e. $K_{ij}^f-Q_i^f$), which can be formulated as,
\vspace{-2mm}
\begin{equation}\label{eq:pointtransformerv2}
\mathcal{G}_{i}=\sum_{j=1}^{K}\mathcal{A}\big(\textcolor{black}{\delta_m} (K_{ij}^f \ominus Q_i^f) \oplus \delta_b \big) \odot \big(V_{ij}^f \oplus \delta_b \big), 
\vspace{-2mm}
\end{equation}

\noindent where $\delta_m = \delta_m (\Delta p_{ij})$ is the position encoding multiplier. %The diagram of the above two classical point transformer are illustrated as Fig.~\hyperref[fig:threepointtransformer]{1.(a)(b)}.  
The attention function $\mathcal{A}(\cdot)$ learns robust weights from the rich relationships, including the low-level geometric relationship (i.e. relation position) and the high-level contextual relationship (i.e. relation feature). \textcolor{black}{Due to the inherent unstructured nature of point clouds, local neighborhoods are typically constructed based on purely position and feature, However, this approach is agnostic to semantic boundaries and often leads to the incorrect aggregation of points from semantically distinct regions.}

%%%%%%%%%%%%%%%%%%%%%%%%%%%%%%%%%%%%%%%%%%%%%%%%%%%%%%%%%%%%%%%%%%%%%%
%%%%%%%%%%%%%%%%%%%%%%%Soft Masked Convolution%%%%%%%%%%%%%%%%%%%%%%%%%
%%%%%%%%%%%%%%%%%%%%%%%%%%%%%%%%%%%%%%%%%%%%%%%%%%%%%%%%%%%%%%%%%%%%%%
% \begin{figure}[t]
% \centering
% \includegraphics[width=  \columnwidth]{softmask.pdf}
% \caption{Illustration of soft mask.}
% \label{fig:softmask}
% \end{figure}

\begin{figure*}[tbp]
\centering
\includegraphics[width= \textwidth]{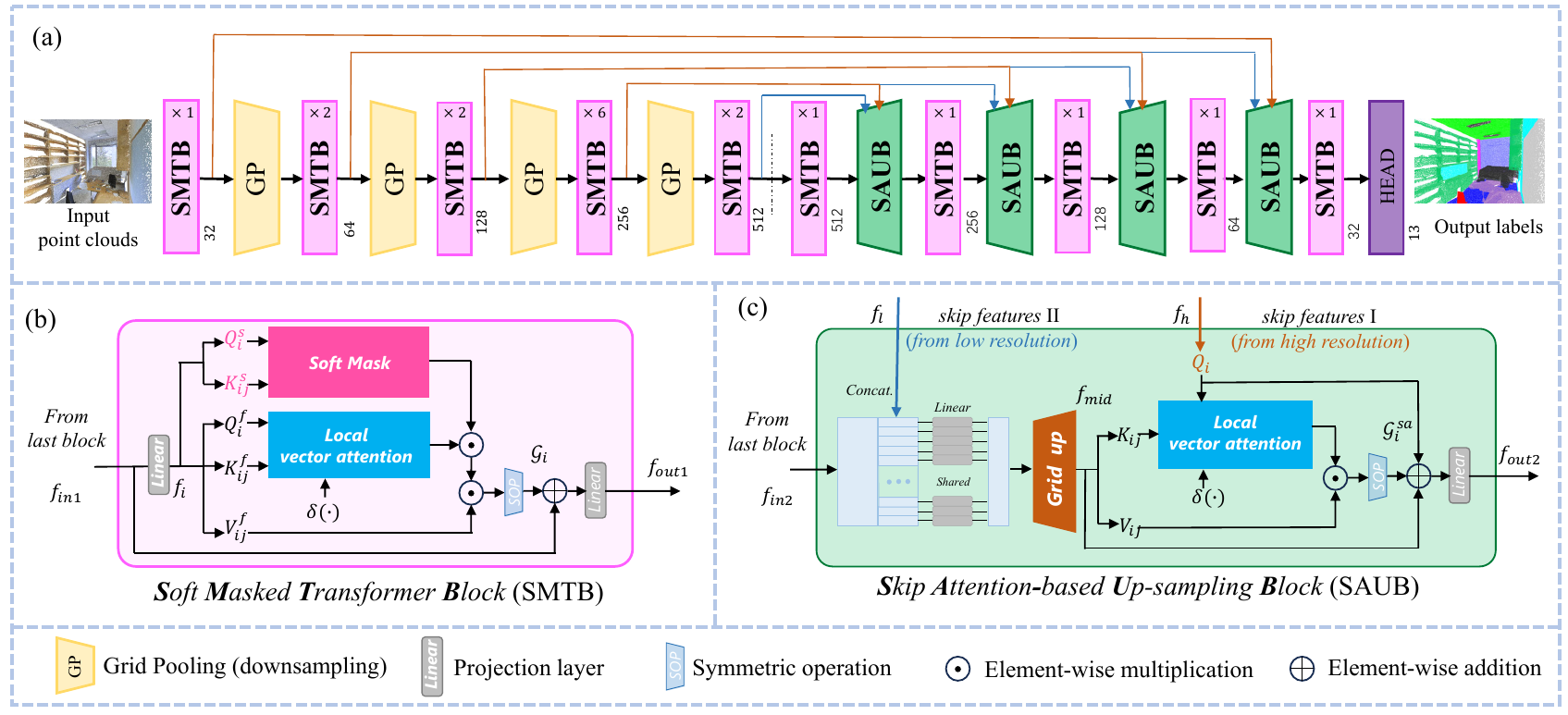}
\caption{ (a) Network architecture for semantic segmentation. (b) Soft masked transformer block and (c) skip attention-based up-sampling block.}
\vspace{-2mm}
\label{fig:archi}
\end{figure*}

\vspace{-2mm}
\subsection{Soft Masked Transformer Block (SMTB)}\label{section3_subsection2}

\textcolor{black}{To address above problem arising from the point cloud data structure, we designed the soft mask transformer. By introducing a learnable semantic signal, it dynamically adjusts attention weights within these geometric neighborhoods, enabling the network to perceive and respect semantic boundaries.} The soft-masked transformer can be expressed as,
\begin{equation}\label{eq:smpointtransformer}
\begin{aligned}
\mathcal{G}_{i} &= {\rm SMTransformer}(f_i),\\
&=\sum_{j=1}^{K}\textcolor{black}{\mathcal{S}(\cdot)} \odot \mathcal{A}\big((K_{ij}^f \ominus Q_i^f) \oplus \textcolor{black}{\delta(\cdot)} \big) \odot \big(V_{ij}^f \oplus \textcolor{black}{\delta(\cdot)}\big), 
\end{aligned}
\end{equation}

\noindent where 
%$K_{ij}^f$, $Q_i^f$ and $V_{ij}^f$ are the feature key, query and value. $\mathcal{A}$ is the attention weight function, implemented by MLPs followed with a softmax function. 
\textcolor{black}{$\mathcal{S}
(\cdot)$} is the soft mask function, which re-weights the attention weights. \textcolor{black}{$\delta(\cdot)$} is the enhanced position encoding function, learning more complex relationships between points. We introduce them in detail in the following text.

\subsubsection{Soft Mask} The soft mask can be interpreted as the learnable coefficient of the attention function. Its significance lies in modelling the semantic context, a prior for calculating a task score difference. This difference is then used to softly mask the attention weights at the point level rather than the channel level. Inspired by the vector attention, SMTransformer divides the features $f_i$ into two entities: the scoring query $Q_i^s$ and score key $K_{ij}^s$, separately.
\begin{equation}
Q_i^s = W_q^s(f_i), ~~ K_{ij}^s = G[W_k^s(f_i)], 
\end{equation}
\textcolor{black}{where the $W_q^s$ and $W_k^s$ ($\mathbb{R}^{N \times C}\rightarrow\mathbb{R}^{N \times T}$) are learnable weight matrices of a shared linear layer that projects the $C$-dimensional feature of each point into a $T$-dimensional task score vector, where $T$ is the number of the classes. These matrices are implemented as 1x1 convolution layers with weight shapes of (C, T) and are optimized end-to-end via backpropagation. The output is passed through a Softmax function along the T-dimension to produce class probabilities.}  $G[\cdot]$ is the grouping operation to obtain the task scores of neighbour points at point $p_i$. The soft mask is generated from the score difference $(K_{ij}^s \ominus Q_i^s) \in\mathbb{R}^{{N\times K \times T}}$ as,
\begin{equation}
S(\cdot) = S(Q_i^s, K_{ij}^s) = || {\rm Max}({\rm Norm}(K_{ij}^s \ominus Q_i^s)) ||_2, 
\end{equation}
where the ${\rm Max}(\cdot)$ is the maximum, ${\rm Norm}(\cdot)$ is Min-Max Normalization, and $||\cdot||_2$ is the Euclidean Norm. 

\textcolor{black}{The resulting soft mask $S(\cdot)$ modulates the original attention weights. A high mask value indicates a strong semantic boundary and is used to down-weight attention to that neighbor, preventing feature corruption from adjacent but semantically distinct regions. This enhances robustness at class boundaries. The module's design is inherently general, as it relies on a learnable task signal (class scores), not on dataset-specific geometry. It can serve as a general-purpose component for point cloud processing. }
% Taking segmentation as an example, the soft mask enhances the robustness of attention weights around class boundaries. Unlike traditional hard masks (i.e., binary masks), soft masks are more flexible and efficient as they do not require explicit rules or conditions for determination.

\subsubsection{Enhanced Position Encoding}
Most existing position encoding methods in local point transformers focus solely on local positions. While this approach greatly assists the transformer in understanding local shapes, it struggles to capture long-range shapes beyond the limited local receptive field. Therefore, global position encoding is equally important as local position encoding. Similarly, SMTransformer encodes the global point position into two entities: position query $Q_i^p$ and position key $K_{ij}^p$.
\begin{equation}
Q_i^p = W_q^p(p_i), ~~ K_{ij}^p = G[Q_i^p], 
\end{equation}
\noindent where the $W_q^p$ ($\mathbb{R}^{N \times 3}\rightarrow\mathbb{R}^{N \times C}$) are global position encoding functions, implemented as MLPs. The local relative position information can be expressed as $ (K_{ij}^p - Q_{i}^p) \oslash \Delta p_{ij}$, where the $\oslash$ denotes the concatenation operation. The enhanced position encoding can be expressed as,
\begin{equation}
\delta(\cdot) = \delta \big( (K_{ij}^p - Q_{i}^p) \oslash \Delta p_{ij}\big),
\end{equation}
\noindent where the $\delta$ is the local position encoding function implemented by MLPs. By constructing the Query and Key matrices of global point position, SMTransformer can bypass the local receptive field limitation to learn the global geometric information.

Residual connections are instrumental in training deep neural networks, facilitating gradient flow during backpropagation. Therefore, we combine the Soft Masked Transformer with residual connections to construct a transformer block. As illustrated in Fig. \ref{fig:archi} (b), the Soft Masked Transformer Block (SMTB) can be expressed as,
\begin{equation}
\begin{aligned}
\vspace{-2mm}
f_i &= {\rm Linear}(f_{in1}),\\
\mathcal{G}_{i} &= {\rm SMTransformer}(f_i),\\
f_{out1} &= {\rm Linear} (\mathcal{G}_{i} \oplus f_i),
\end{aligned}
\end{equation}
where $f_{in1}$ is the input feature and $f_{out1}$ is the output feature of SMTB. The projection layer $\rm Linear()$ is achieved through a series of layers, including one linear layer, one batch normalization layer, and one Relu layer.

\begin{figure}[tbp]
\centering
\includegraphics[width=  \columnwidth]{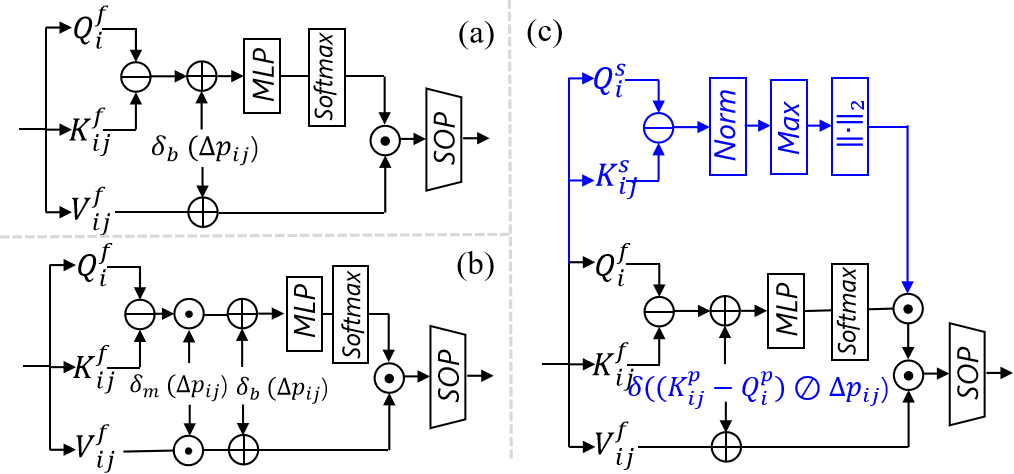}
\caption{ Comparison of the attention, position encoding in  Transformers. (a) The vector attention with position encoding bias in Point Transformer, see Eq.\eqref{eq:pointtransformer}. (b) The vector attention with position encoding multiplier in Point Transformer V2, see Eq.\eqref{eq:pointtransformerv2}. (c) The vector attention with soft mask and enhanced position encoding bias in our proposed SMTransformer, see Eq.\eqref{eq:smpointtransformer}. \textcolor{black}{The different parts are coloured in blue}.}
\vspace{-4mm}
\label{fig:threepointtransformer}. 
\end{figure}

To compare the differences between SMTransformer and the classical vector attention-based point transformer, we illustrate their architectures in Fig. \ref{fig:threepointtransformer}. There are two key distinctions:

\textbf{i)} Both Point Transformer and Point Transformer V2 emphasize learning contextual relationships. In contrast, SMTransformer not only grasps contextual relationships through vector attention but also integrates a soft mask as the coefficient with the attention function, driven from the task at hand.

\textbf{ii)} Point Transformer and Point Transformer V2 both effectively capture local fine-grained details through local position encoding. In contrast, SMTransformer introduces an innovative enhanced position encoding that represents positions across the global point cloud, enabling modelling of the global shape without being confined to local receptive fields. Additionally, it encodes positions across local points, allowing for learning fine-grained details.

\vspace{-2mm}
\subsection{Skip Attention-based Up-sampling Block (SAUB)}\label{section3_subsection3}

\begin{comment}
    We propose a skip attention-based upsampling block that combines the conventional up-sampling (i.e. grid unspooling) with one learnable unit (i.e. skip attention) to learn and refine the contextual information between the feature from the encoding layer and decoding layers over different resolution points. As illustrated in Fig.\ref{fig:archi}(c). Given the input feature and skip feature II from the encoding layer over low-resolution points, the SAUB first concatenates them.
\end{comment}

\textcolor{black}{Conventional upsampling often use simplistic upsampling schemes (e.g., interpolation) that are incapable of effectively recovering the fine-grained geometric details lost during upsampling.} To address this problem, we introduce a skip attention-based up-sampling block that combines conventional unpooling with a learnable unit to learn and refine contextual information between features from the encoding and decoding layers across different resolutions, \textcolor{black}{facilitating deep communication between features over various resolution points.}

 %the input and output features of block are $f_{in2}$ and $f_{out2}$

As illustrated in Fig. \ref{fig:archi} (c), given the skip feature I $f_h\in\mathbb{R}^{{M\times C_h}}$ and skip feature II $f_l\in\mathbb{R}^{{m\times C_l}}$ from the two adjacent encoding layers over the low and high-resolution points, respectively. To build the communication between the different resolution point features, we first balance their feature dimension and point resolution,
\begin{equation}
\vspace{-2mm}
f_{mid} = {\rm Grid up}\big({\rm Linear}(f_{in2} \oslash f_l)\big),
\end{equation}
where the ${\rm Linear()}$ serves as the projection layer; its primary role involves integrating both the input features $f_{in2}$ and skip features II $f_l$, thereby augmenting the dimension of low-resolution point features to match that of high-resolution point features ($\mathbb{R}^{{m\times C_l}} \rightarrow \mathbb{R}^{{m\times C_h}}$), where $m$ is the number of low-resolution points. The ${\rm Grid up}$ is the common practice of unpooling, implemented by grid-based unpooling, to augment the resolution ($\mathbb{R}^{{m\times C_h}} \rightarrow \mathbb{R}^{{M\times C_h}}$), where $M$ is the number of high-resolution points. By the above two steps, the low-resolution features are expanded in terms of feature dimension and point resolution.

To build deep communication between the expanded low-resolution features and skip features I, we propose the skip attention, 
\begin{equation}
Q_i = w_q(f_l), K_{ij} = G[w_k(f_{mid})], V_{ij} = G[w_v(f_{mid})],
\end{equation}
\begin{equation}
\vspace{-2mm}
\mathcal{G}_{i}^{sa} =\sum_{j=1}^{K}\mathcal{A}\big((K_{ij} \ominus Q_i) \oplus \textcolor{black}{\delta(\cdot)}\big) \odot \big(V_{ij} \oplus \textcolor{black}{\delta(\cdot)}\big), 
\end{equation}
where $\delta(\cdot)$ is the proposed enhanced position encoding. $\mathcal{G}_{i}^{sa}$ is the output of skip attention. The skip attention use $Q_i$, derived from high-resolution features $f_l$, as query and $K_{ij}$, derived from expanded low-resolution features $f_{mid}$ to learn the contextual connection (i.e. attention map) between different resolution points.

To prevent the loss of some important information, we utilize a residual connection, concentrating the output on the input of skip attention and skip features, 
\begin{equation}
f_{out2} = {\rm Linear}(\mathcal{G}_{i}^{sa}  \oplus f_l  \oplus f_{mid}).
\end{equation}

\begin{figure}[tbp]
\centering
\includegraphics[width=  \columnwidth]{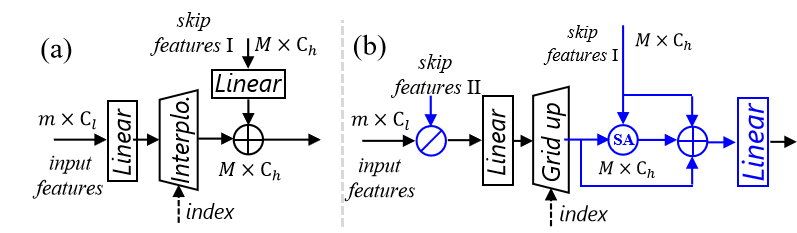}
\vspace{-4mm}
\caption{ Comparison of (a) Transition up and (b) Skip attention-based up-sampling. `interpo.' stands for the interpolation operation. `grid up' stands for the grid-based unpooling. `SA' denotes the skip attention. \textcolor{black}{The different parts are coloured in blue.}}
% \vspace{-2mm}
\label{fig: skip_attention}
\end{figure}

\begin{comment}
  To compare the differences between SAUB and classical grid unspooling, we illustrate their diagram in Figure \ref{fig: skip_attention}. The key differences lie in two aspects: \textbf{1)} the classical grid unspooling method just learns the connection between features from the encoding and decoding layer over the same resolution points. Still, our proposed SAUB could learn the connection between features from encoding and decoding layers over the different resolution points. \textbf{2)} The grid unspooling uses the simple skip connection to learn the contextual information between the features from the encoding and decoding layer, but we propose the skip attention to refine the contextual information between the features.  
\end{comment}
To compare the differences between SAUB and Conventional upsampling block ( i.e. Transition up\cite{wu2022point}), we illustrate their diagrams in Fig. \ref{fig: skip_attention}. 
\textcolor{black}{ The Skip Attention-Based Upsampling (SAUB) module to facilitate a more intelligent fusion of multi-resolution features. Instead of relying on simple unpooling, SAUB incorporates a learnable attention mechanism that actively learns and refines the contextual information exchanged between encoder and decoder features. This allows it to dynamically adapt to the local point structure, ensuring a detail-preserving and context-aware upsampling process.}

% The key differences lie in two aspects:

% \textbf{i)} The classical upsampling block only learns the connection between features from the encoding and decoding layers over the same resolution points. In contrast, our proposed SAUB can learn the connection between features from the encoding and decoding layers of different resolution points.

% \textbf{ii)} The classical upsampling block uses a simple skip connection to learn the contextual information between the features from the encoding and decoding layers. However, we propose \textit{skip attention} to refine the contextual information between the features.

\begin{figure}[h]
\centering
\includegraphics[width=3.2in]{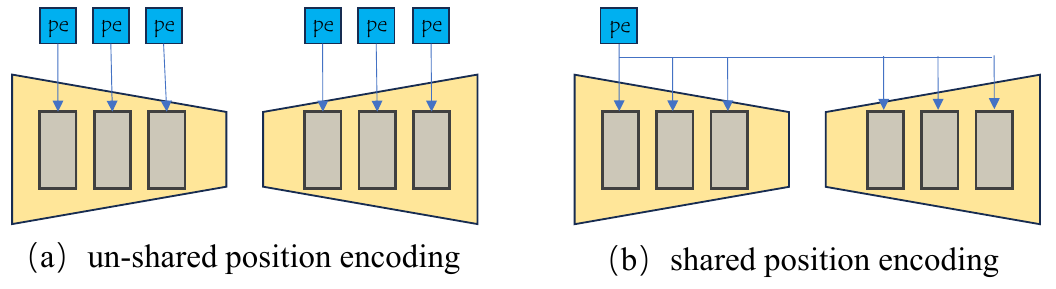}
\vspace{-2mm}\caption{(a) Unshared point position encoding: Various transformer blocks (coloured in grey) within the same encoding or decoding layer (coloured in yellow), operating over the same resolution point cloud, require different position information (coloured in blue). (b) Shared point position encoding: Various transformer blocks within the encoding and decoding layers share position information over the same resolution point cloud. } 
\label{fig: spe}
\end{figure}

\vspace{-4mm}
\subsection{Shared Point Position Encoding}\label{section3_subsection4}
In conventional point transformer networks, several transformer blocks typically utilize different position encoding information within the same encoding or decoding layer across resolution points. We refer to this practice as unshared point position encoding, \textcolor{black}{which would increase the network parameter and training time.} Intuitively, points within the same position in the point cloud should have the same position information. Building on this intuition, we propose the shared point position encoding strategy. Under this strategy, transformer blocks within the same encoding or decoding layer share identical position encoding information across resolution points. This approach enhances network robustness and efficiency, particularly in large-scale scene processing. Their comparison is illustrated in Fig. \ref{fig: spe}. 
% \textcolor{black}{Besides, SPE could decrease the parameters and computational cost of the whole network.}
% The robustness and efficiency experiments \textcolor{black}{are provided} in Section \ref{section4_subsection4}.

\vspace{-2mm}
\subsection{Network Architecture}\label{section3_subsection5} 

\begin{comment}
    For the semantic segmentation task, we use a U-Net-like architecture with five encoding and decoding layers and skip connections. The first encoding and decoding layers consist of one MLP and SMTransformer block. Other encoding layers contain one Grid Pooling\cite{wu2022point} and several SMTransformer blocks, where the number of SMTransformer blocks in the five encoding layers is [1, 2, 2, 6, 2]. The decoding layers consist of one skip attention-based up-sampling. We set the feature dimensions $C$ as [32, 64, 128, 256, 512] for the five encoding and decoding layers. At the end of the network, we add an MLP to predict the final point-wise labels. The network architecture on S3DIS is illustrated in Fig.\ref{fig:archi}. For the classification task, we use the basic PointNext\cite{qian2022pointnext} as the backbone and replace the one SMTransformer block with the MLPs block to form the new network architecture for classification. The more detailed segmentation and classification network configurations will be introduced in Section \ref{section 4}.

\end{comment}

We employ a U-Net-like architecture comprising five encoding and decoding layers with skip connections for the semantic segmentation task. The first encoding and decoding layers consist of one MLP and SMTransformer block. The subsequent encoding layers incorporate one Grid Pooling layer \cite{wu2022point} followed by several SMTransformer blocks. The number of SMTransformer blocks in the five encoding layers is [1, 2, 2, 6, 2]. The \textcolor{black}{subsequent} decoding layers consist of one skip attention-based up-sampling block \textcolor{black}{ and one SMTransformer block}. We set the feature dimensions as [32, 64, 128, 256, 512] for the 5 encoding and decoding layers. At the network's end, we append an MLP to predict the final point-wise labels. The network architecture for %the S3DIS dataset 
semantic segmentation is illustrated in Fig. \ref{fig:archi} (a).

% For the classification task, we utilize the basic PointNext \cite{qian2022pointnext} as the backbone and replace one SMTransformer block with an MLPs block to form the new network architecture. Further details regarding the configurations of the segmentation and classification networks are in Section \ref{section4}.

% \vspace{-2mm}
\section{Experimental Results}\label{section4}
\textcolor{black}{We evaluate our network on two tasks namely, semantic segmentation of indoor scenes semantic segmentation and outdoor scene semantic segmentation. We also perform detailed ablation studies to demonstrate the effectiveness and robustness of the proposed Soft Mask Transformer Block, Skip Attention-based up-sampling Block and the Shared Point Position Encoding strategy.}

\vspace{2mm}
\begin{figure*}[t]
\centering
\includegraphics[width= 0.98 \textwidth]{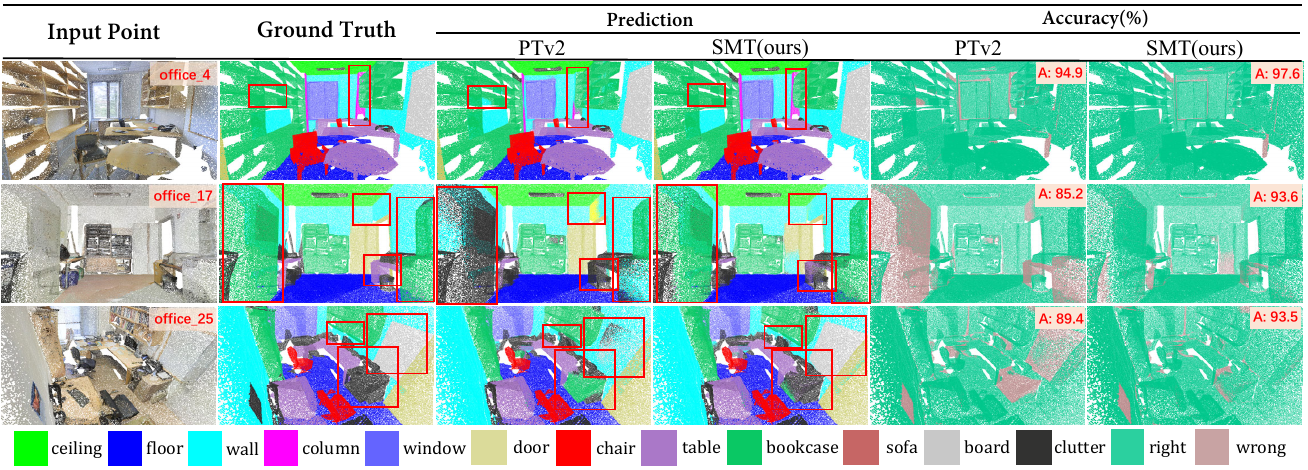}
% \vspace{-2mm}
\caption{Visualization of semantic segmentation results on S3DIS Area-5. The \textcolor{black}{red} boxes highlight the object boundaries in the scenes where our proposed SMTransformer performs particularly better than the Point Transformer V2 (PTv2).}
\vspace{-4mm}
\label{fig:s3disVis}
\end{figure*}

\vspace{-4mm}
\subsection{Indoor Semantic Segmentation}

\begin{comment}
    We evaluate our network on two large-scale indoor scene datasets, S3DIS\cite{armeni20163d} and ScanNetV2\cite{dai2017scannet}. The S3DIS  contains RGB-D point clouds annotated point-wise with 13 classes. It covers 271 rooms from 6 large-scale indoor scenes (a total of 6020 square meters). We use the 6-dimensional point features, including three dimensions normalized colour and 3 dimensions normalized location. We conduct 6-fold cross-validation on S3DIS and, in line with other works, conduct more extensive comparisons on Area 5 as the test set, which is not in the same building as the other areas. The ScanNetV2 contains coloured point clouds of indoor scenes with point-wise semantic labels of 20 object categories. It is split into 1201 scenes for training and 312 for validation.  We use the 9-dimensional point features including three dimension normalized color, 3 dimension normalized location and 3 dimension normal.
\end{comment}
\textit{Datasets:} We evaluate our network on two large-scale indoor scene datasets, namely S3DIS \cite{armeni20163d} and ScanNetV2 \cite{dai2017scannet}. The S3DIS dataset consists of RGB-D point clouds annotated point-wise with 13 classes. It encompasses 271 rooms from 6 large-scale indoor scenes, totalling 6020 square meters. We utilize 6-dimensional point features, including 3-dimensional normalized colour and 3-dimensional normalized location. For evaluation, we focus more extensively on comparisons using Area 5 as the test set, which is distinct from the other areas and not within the same building.

The ScanNetV2 dataset comprises coloured point clouds of indoor scenes with point-wise semantic labels for 20 object categories. It is divided into 1201 scenes for training and 312 for validation. Our approach utilizes 9-dimensional point features corresponding to 3-dimensional normalized colour, 3-dimensional normalized location, and 3-dimensional normals.
\vspace{0.5mm}
\textit{Network Configurations:} 
For semantic segmentation on S3DIS,  we set the voxel size as 4cm and the maximum number of voxels to 60,000. We adopt the SGD optimizer and weight decay as 0.0001. The base learning rate is set as 0.4 and the learning rate is scheduled by the MultiStepLR at the 40\textit{th} and 80\textit{th} epoch. We train and test the model with batch size 16 and 8 on 4 GPUs, respectively. We adopt random scaling, random flip,  chromatic contrast, chromatic translation, chromatic jitter and hue saturation translation to augment training data. We set the grid size in grid pooling as [0.08, 0.1, 0.2, 0.4]\textit{cm} and the number of neighbour points in SMTransformer as 16.

On ScanNetV2, we set the voxel size as 2cm and the maximum number of voxels to 100,000. We use the Adam optimizer, where the weight decay is set as 0.02. The base learning rate is set as 0.02 and the learning rate is scheduled by the MultiStepLR every 40 epochs. We train and test the model with batch size 24 on 4 GPUs. We adopt random rotation, random scaling, random flip, elastic distortion, chromatic contrast, chromatic translation, chromatic jitter and hue saturation translation to augment training data. The grid size is set as [0.04, 0.12, 0.36, 1.08]\textit{cm}, and the number of neighbour points is set as 16. During the test, the network uses the test time augmentation, following the Point Transformer V2\cite{wu2022point} and Stratified Transformer\cite{lai2022stratified}. 

\begin{table}[tbp]
% \vspace{-2mm}
\centering
\caption{Semantic segmentation results on the S3DIS dataset Area-5. We report mean class-wise Intersection over Union (mIoU), mean class-wise accuracy (mAcc), overall accuracy (OA), and parameter (P). \textit{w}/T denotes the model test with test time augmentation trick. The best result is highlighted in \textbf{bold}. The second best result is \underline{underlined} }
\vspace{-2mm}
\label{table: s3dis area5}
\scriptsize
\renewcommand\arraystretch{1.2}
\setlength{\tabcolsep}{1.2mm}{
\begin{tabular}{l|l|c|ccc|c}
\specialrule{1pt}{0pt}{0pt}
{Year} &{Methods}    &\textit{w}/T  & {mIoU}   & {mAcc}  & {OA}   &Para.(M)  \\\hline
% 2017 CVPR &PointNet\cite{qi2017pointnet} &$\times$ & 41.09  & 48.98 &--   & \underline{3.6 }  \\
% 2018 NIPS &PointCNN\cite{li2018pointcnn}&$\times$ & 57.26    & 63.86 &85.9    &\textbf{0.6} \\
% 2019 ICCV&KPConv\cite{thomas2019kpconv}&$\times$ & 67.1     & 72.8 &--    &15    \\
% 2020 PAMI &SPH3D-GCN\cite{lei2020spherical}&$\times$ & 59.5 & 65.9 & --   &--   \\

% 2020 CVPR &PointANSL\cite{yan2020pointasnl}&$\times$  &62.6 &68.5 &87.7  &22.4   \\
% 2020 CVPR &SegGCN\cite{Lei_2020_CVPR} &$\times$ &63.6   &70.4 &--  &--   \\
% 2021 CVPR &PAConv\cite{xu2021paconv}&$\times$ & 66.6 & 73.0   &--   &--     \\
% 2021 CVPR &BAAF-Net\cite{qiu2021semantic} &65.4 &73.1 &88.9 &$\times$&   &72.2 &{83.1} &88.9   &$\times$   \\
% 2022 CVPR &CBL\cite{tang2022contrastive}  &$\times$&69.4  &75.2 &90.6  & 18.6       \\
% 2022 CVPR &RepSurf-U\cite{ran2022surface} &$\times$&68.9 &76.0 & 90.2  &\textbf{1}  \\
% 2022 ECCV &PointMixer\cite{choe2022pointmixer} &$\times$ &71.4 &77.4   &--  &6.5 \\
% 2022 NIPS &PointNeXt\cite{qian2022pointnext} &$\times$&70.5 &76.8 &90.6  &41.6    \\
% 2022 TMM &3D-PAM\cite{weng2022context} &$\times$&68.4 &-- &--  &-- \\
% 2022 TITS &3DCTN\cite{lu20223dctn} &$\times$&72.1 &-- &\textbf{92.5}  &-- \\
% 2023 CVPR &Point Vector\cite{deng2023pointvector}&$\times$ &{72.3} &{78.1} &91.0  &24.1\\
% 2023 TNNLS &PicassoNet++\cite{lei2023mesh}&$\times$ &71.0 &77.2 &91.3   &-- \\
\textcolor{black}{2024 TMM} &\textcolor{black}{EPNet\cite{deng2024ep}} &\textcolor{black}{$\times$} &\textcolor{black}{72.8} &\textcolor{black}{78.3} &\textcolor{black}{91.5} &\textcolor{black}{14.8}   \\

\textcolor{black}{2024 CVPR} &\textcolor{black}{OneFormer3D\cite{kolodiazhnyi2024oneformer3d}} &\textcolor{black}{$\times$}  &\textcolor{black}{72.4} &\textcolor{black}{--} &\textcolor{black}{--} &\textcolor{black}{--}\\

\textcolor{black}{2024 CVPR} &\textcolor{black}{KPConvX-L\cite{thomas2024kpconvx}} &\textcolor{black}{$\times$}  &\textcolor{black}{72.3} &\textcolor{black}{78.4} &\textcolor{black}{91.4} &\textcolor{black}{--}\\

\textcolor{black}{2025 AAAI} &\textcolor{black}{DeepLA-24\cite{zeng2025deepla}} &\textcolor{black}{$\times$}  &\textcolor{black}{73.2} &\textcolor{black}{--} &\textcolor{black}{--} &\textcolor{black}{--}\\

\textcolor{black}{2025 IJAEOG} &\textcolor{black}{U-Next\cite{zeng2025small}} &\textcolor{black}{$\times$}  &\textcolor{black}{72.7} &\textcolor{black}{--} &\textcolor{black}{91.2} &\textcolor{black}{--}\\

\textcolor{black}{2025 IEEE ITJ} &\textcolor{black}{BeyondPoints\cite{luo2025beyondpoints}} &\textcolor{black}{$\times$}  &\textcolor{black}{73.1} &\textcolor{black}{79.1} &\textcolor{black}{91.2} &\textcolor{black}{--}\\

\textcolor{black}{2025 AAAI} &\textcolor{black}{Point Manba\cite{zhang2025point}} &\textcolor{black}{$\times$}  &\textcolor{black}{\bf 79.6} &\textcolor{black}{--} &\textcolor{black}{--} &\textcolor{black}{--}\\

% 2024 TMM &\\

% 2024 TMM &MPCT\cite{wu2023mpct} &$\times$ &{68.6} &{74.6} &89.3 &--\\
% 2024 TMM &PointGL\cite{li2024pointgl} &$\times$ &{65.6} &{71.9} &88.6 &3.5\\
\hline

\multicolumn{7}{c}{\textit{previous state-of-the-art point transformer}\textcolor{orange}{$\star$}} \\\hline

2021 ICCV &Point Transformer\cite{zhao2021point}&$\times$&70.4 &76.5 &90.8  & \underline{7.8}  \\
2022 CVPR &Stratified Transformer\cite{lai2022stratified} &$\times$&{71.4} &{77.4} & {90.9}  & 18.8 \\
2022 NIPS &Point Transformer V2\cite{wu2022point} &$\times$&70.7 &76.3 &{90.9}  & 11.3  \\
2023 NIPS &ConDAForMer\cite{duan2024condaformer}  &$\times$  &{72.2} &{78.2} &{91.2}   &8.4 \\
2024 CVPR &Point Transformer V3\cite{wu2023point} &$\times$ &72.0 &77.8 &{91.0}   &46.2  \\
2024 TNNLS &Full Point Transformer\cite{he2023full} &$\times$ &72.2 &{78.5} &{91.5}   &10.9\\
&\textbf{SMTransformer}(ours) &$\times$ &{72.6} &{78.8} &{91.7}  &\underline{7.8}  \\\hline

2021 ICCV &Point Transformer\cite{zhao2021point} &$\checkmark$&70.8 &77.0 &91.0   & \underline{7.8} \\
2022 CVPR &Stratified Transformer\cite{lai2022stratified}&$\checkmark$ &{71.9} &{78.0} & {91.2}   &18.8 \\
2022 NIPS &Point Transformer V2\cite{wu2022point} &$\checkmark$&71.6 &77.5 &{91.0}  &11.3 \\
2023 NIPS &ConDAForMer\cite{duan2024condaformer}&$\checkmark$ &{73.2} &{78.4} &{91.2}  &8.4 \\
2024 CVPR &Point Transformer V3\cite{wu2023point} &$\checkmark$ &73.1 &78.3 &{91.1}  &46.2\\
2024 TNNLS &Full Point Transformer\cite{he2023full} &$\checkmark$ &{73.1} &{78.8} &{91.7}  &{10.9}  \\
&\textbf{SMTransformer}(ours) &$\checkmark$ &{73.6} &\underline{79.0} &\underline{91.8}  &\underline{7.8}\\
&\textbf{SMTransformer}\textcolor{orange}{$\dagger$} (ours) &$\checkmark$ &\underline{73.9} &\textbf{79.3} &\textbf{93.0}  &{24.5}\\\hline
\multicolumn{7}{l}{\textcolor{black}{\textcolor{orange}{$\star$}: reproduced the original code by the same workstation.}}\\
\multicolumn{7}{l}{\textcolor{black}{\textcolor{orange}{$\dagger$}: denotes the method uses multi-scale semantic information.}}\\
\multicolumn{7}{l}{\textcolor{black}{All methods do not use multi-dataset joint training}}\\
\end{tabular}}
\vspace{-4mm}
\end{table}

\begin{table}[t]
\centering
% \vspace{-2mm}
\caption{Semantic segmentation results (mIoU) on ScanNetV2 validation set.}
\vspace{-2mm}
\label{table: scannet sementic}
\scriptsize
\renewcommand\arraystretch{1.2}
\setlength{\tabcolsep}{1.4mm}{
\begin{tabular}{l|l|c|c|c}
\specialrule{1pt}{0pt}{0pt}
{Year} &{Methods}     &\textit{w}/T & mIoU(\%)   &Para.(M) \\ \hline
% 2018 NIPS &PointNet++\cite{qi2017pointnet++}  &$\times$ &\textcolor{black}{55.7} &--\\
% 2018 CVPR &SparseConvNet\cite{graham20183d}   &$\times$ &\textcolor{black}{72.5} &37.9   \\
% 2019 CVPR &PointConv\cite{wu2019pointconv}  &$\times$ &\textcolor{black}{66.6}   &9.3 \\
% 2020 CVPR &PointANSL\cite{yan2020pointasnl}   &$\times$ &63.5 &22.4\\
% 2019 ICCV &KPConv\cite{thomas2019kpconv}   &$\times$ &69.2  &15 \\
% 2019 3DV &JointPointBased\cite{chiang2019unified} &$\times$ &69.2 &-- \\
% 2019 CVPR &\textcolor{black}{MinkowskiNet}\cite{choy20194d}  &$\times$ &72.2  & 37.9\\
% 2022 TMM &3D-PAM\cite{weng2022context} &$\times$&70.6  &-- \\
% 2022 CVPR &RepSurf-U\cite{ran2022surface} &$\times$&70.0  &\textbf{1} \\
% 2022 NIPS  &PointNeXt\cite{qian2022pointnext} & $\times$ &71.5 &41.6 \\
% 2022 CVPR &Fast Point Transformer\cite{park2022fast}  &$\times$ &72.4  &37.9 \\
% 2023 CVPR &LargeKernel3D\cite{chen2023largekernel3d} &$\times$&73.5 &40.2 \\
% 2023 CVPR &PointMetaBase\cite{lin2023meta} &$\times$&72.8  &19.7 \\
% 2023 CVPR  &PointConvFormer\cite{wu2023pointconvformer} &$\times$&74.5 &9.4 \\

\textcolor{black}{2024 CVPR} &\textcolor{black}{OneFormer3D\cite{kolodiazhnyi2024oneformer3d}} &\textcolor{black}{$\times$}  &\textcolor{black}{\bf 76.6} &\textcolor{black}{--}\\

\textcolor{black}{2024 CVPR} &\textcolor{black}{KPConvX-L\cite{thomas2024kpconvx}} &\textcolor{black}{$\times$}  &\textcolor{black}{75.4} &\textcolor{black}{--}\\

\textcolor{black}{2025 AAAI} &\textcolor{black}{DeepLA-24\cite{zeng2025deepla}} &\textcolor{black}{$\times$}  &\textcolor{black}{74.1} &\textcolor{black}{--}\\

\textcolor{black}{2025 IEEE ITJ} &\textcolor{black}{BeyondPoints\cite{luo2025beyondpoints}} &\textcolor{black}{$\times$}  &\textcolor{black}{74.6} &\textcolor{black}{--}\\

\textcolor{black}{2025 AAAI} &\textcolor{black}{Point Manba\cite{zhang2025point}} &\textcolor{black}{$\times$}  &\textcolor{black}{75.5} &\textcolor{black}{--}\\
% 2023 TMM 
\hline

\multicolumn{5}{c}{\textit{previous state-of-the-art point transformer}\textcolor{orange}{$\star$}}\\\hline
2022 CVPR &Stratified Transformer\cite{lai2022stratified}  &$\times$&73.0 &18.8\\ 
2022 NIPS & Point Transformer V2\cite{wu2022point}  &$\times$ & 74.5 &12.8\\
2023 NIPS & ConDAForMer\cite{duan2024condaformer}  &$\times$  & 75.2 &8.4\\
2024 CVPR &Point Transformer V3\cite{wu2023point}   &$\times$ & {76.2} &46.2\\
2024 TNNLS &Full Point Transformer\cite{he2023full} &$\times$ &74.8   &10.9\\
&\textcolor{black}{\textbf{SMTransformer} (ours)}  &$\times$ &{75.9} &\underline{7.8} \\\hline

2022 CVPR &Stratified Transformer\cite{lai2022stratified}  &$\checkmark$&73.8 &18.8 \\ 
2022 NIPS & Point Transformer V2\cite{wu2022point}  &$\checkmark$ & 75.0 &12.8 \\
2023 NIPS & ConDAForMer\cite{duan2024condaformer}  &$\checkmark$  & 75.6 & 8.4\\
2024 CVPR &Point Transformer V3\cite{wu2023point}  &$\checkmark$ & \textbf{76.6} &46.2\\
2024 TNNLS &Full Point Transformer\cite{he2023full} &$\checkmark$ &75.6   &10.9\\
&\textcolor{black}{\textbf{SMTransformer} (ours)} &$\checkmark$ &\underline{76.2} &\underline{7.8} \\\hline
\multicolumn{5}{l}{\textcolor{black}{\textcolor{orange}{$\star$}: reproduced the original code by the same workstation.}}\\
\multicolumn{5}{l}{\textcolor{black}{All methods do not use multi-dataset joint training}}\\
\end{tabular}}
\vspace{-4mm}
\end{table}
% \vspace{0.5mm}

% \vspace{-2mm}
\begin{table*}[tbp]
\caption{Semantic segmentation results on the Swan test set. }%We reproduced these experiments using their original code on the Swan dataset.}
\vspace{-2mm}
\label{table:swan}
\scriptsize
\centering
\renewcommand\arraystretch{1.3}
\setlength{\tabcolsep}{1.mm}{
\begin{tabular}{l|l|c|ccccccccccccccccccccc}
\specialrule{1pt}{0pt}{0pt}
Year &{Methods} &\rotatebox{90}{{mIoU}(\%)} &\rotatebox{90}{car} &\rotatebox{90}{truck} &\rotatebox{90}{pedes.} &\rotatebox{90}{bicycle} &\rotatebox{90}{motor.~~} &\rotatebox{90}{bus} &\rotatebox{90}{bridge} &\rotatebox{90}{tree} &\rotatebox{90}{bushes} &\rotatebox{90}{building} &\rotatebox{90}{road} &\rotatebox{90}{r-driver} &\rotatebox{90}{rub-bin} &\rotatebox{90}{bus-stop} &\rotatebox{90}{pole} &\rotatebox{90}{wall} &\rotatebox{90}{sign} &\rotatebox{90}{rs-board} &\rotatebox{90}{sidewalk} &\rotatebox{90}{adv-board}\\\hline
2017 NIPS &Pointnet++\cite{qi2017pointnet++} &14.5 & 31.2 & 7.3 & 4.7 &9.0 & 0.0 &4.8 & 0.0 & 33.9 &12.7 &59.5 & 68.4 &0.0 &13.7 &9.0 &6.9 &15.9 &1.0 & 2.1 & 11.0 &0.0 \\

2019 CVPR &PointConv\cite{wu2019pointconv} &37.3 &53.7 &20.5 &36.5 &19.6 & 5.2 &68.7 & 7.7 & 61.0 &52.5 &74.2 &77.1 &42.3 &19.6 &38.9 &22.1 &45.1 &26.7 & 27.3 & 40.4 &6.4 \\

2019 CVPR&$\psi$-CNN\cite{lei2019octree} &39.8 &48.5 &25.2 &31.1 &22.4 & 4.2 &77.6 & 7.3 & 69.0 &56.2 &73.9 &75.6 &47.1 &23.0 &46.1 &30.1 &57.0 &29.1 &25.8 &38.2 &9.4 \\

2021 CVPR &PolarNet\cite{zhang2020polarnet} &40.5 &78.1 &20.3 &21.6 &4.6 & 15.3 &18.0 & 8.3 & 84.0 &30.9 &91.9 &92.7 &54.7 &33.4 &29.5 &48.0 &60.3 &42.7 & 22.2 & 42.2 &11.8 \\

2021 CVPR&Cylinder3D\cite{zhu2021cylindrical} &54.9 &80.8 &30.4 &{48.7} &28.0 &6.8 &{91.7} &13.7 &{85.3} &{69.0} &92.7 &{92.3} &{75.2} &37.5 &72.1 &{48.8} & {71.1} & {46.5} & {32.5} & 56.7 &18.9 \\

2021 ICCV &Point Transformer \cite{zhao2021point} 
            & {58.6}

            & {84.5} 
            & {46.2} 
            
            &40.4 
            & {42.0}  
            & {13.3} 
           & {83.5} 
           & {50.3} 
           & {85.2} 
           &  67.6 
           & 93.5 
           &{90.3} 
           &81.7 
           & {51.8} 
           &{75.1} 
           &40.7 
           &{72.0} 
           &47.0 
           &30.5 
           & {61.2} 
           & {15.3} 
           %20   1234.6
           \\

 2022 NIPS&Point Transformer V2 \cite{wu2022point}  & {59.7} & {84.7} & {47.7} &41.2 & {43.1} & {14.3} 
           & {87.5} 
           & {51.3} 
           & {86.3} 
           & 69.2 
           & {92.8} 
           &{91.3} 
           &80.2 
           & {52.6} 
           &{76.0} 
           &41.5 
           &{72.4} 
           &48.5 
           &32.2 
           & {63.4} 
           & {17.2} 
\\
2024 TITS&SAT3D\cite{ibrahim2023sat3d} & {58.2} & {83.7} & {45.7} &38.7 & {42.4} & {11.3} & {89.5} & {49.3} & {85.5} &68.2 & {93.2} &{92.7} &74.9 & {40.6} &{77.1} &43.3 &{74.8} &42.9 &26.7 & {65.6} & {17.9} \\

2024 CVPR &Point Transformer V3 \cite{wu2023point} & \underline{61.1} & {84.7} & {48.2} &45.7 & {43.3}   & {16.3} 
            & {88.5} 
            & {53.2}
            & {87.7} 
            &  69.8 
            & {94.8} 
            &{92.7} 
            &81.4 
            & {53.3} 
            &{77.2} 
            &45.3 
            &{73.8} 
            &50.1
            &33.6 
            & {64.1} 
            & {18.9} 
\\

&{SMTransformer}(ours) &\textbf{62.4} &{86.4} & {50.2} & {47.2} & {44.6} & {19.6} &{87.0} & {56.0}  & {88.4} &{70.4} &{94.0} &92.2 &{83.3} &{56.8} & {75.1} & {46.7} &{74.8} &{54.9} &{35.6} & {67.0} & {21.0} \\\hline
% \multicolumn{5}{l}{$\star$: reproduced the original code on Swan.}\\

\end{tabular}
}
\end{table*}

% Anwaar: Typos are fixed

%%%%%%%%%%%%%%%%semantickiiti%%%%%%%%%%%%%%%%%%%
%%%%%%%%%%%%%%%%%%%%%%%%%%%%%%%%%%%%%%%%%%%%%%%%
% \vspace{-2mm}
\begin{table*}[t]
% \vspace{-2mm}
\caption{\textcolor{black}{Semantic segmentation results on the SemanticKITTI. \textcolor{black}{`*' means the network does not use any extra data.}}}
% \vspace{-2mm}
\label{table: kitti}
\scriptsize
\centering
\renewcommand\arraystretch{1.2}
\setlength{\tabcolsep}{1.mm}{
\begin{tabular}{l|l|c|ccccccccccccccccccc}
\specialrule{1pt}{0pt}{0pt}
{Year} &Methods &\rotatebox{90}{{mIoU}(\%)} &\rotatebox{90}{car} &\rotatebox{90}{bicycle} &\rotatebox{90}{motor.} &\rotatebox{90}{truck} &\rotatebox{90}{other-vehi.} &\rotatebox{90}{person} &\rotatebox{90}{bicyclist} &\rotatebox{90}{motorcyc.~~} &\rotatebox{90}{road} &\rotatebox{90}{parking} &\rotatebox{90}{sidewalk} &\rotatebox{90}{other-gro.} &\rotatebox{90}{building} &\rotatebox{90}{fence} &\rotatebox{90}{vegetation} &\rotatebox{90}{trunk} &\rotatebox{90}{terrain} &\rotatebox{90}{pole} &\rotatebox{90}{sign} \\\hline

% 2017 NIPS &Pointnet++\cite{qi2017pointnet++} &20.1 &53.7 &1.9 &0.2 &0.9 &0.2 &0.9 &1.0 &0.0 &72.0 &18.7 &41.8 &5.6 &62.3 &16.9 &46.5 &13.8 &30.0 &6.0 &8.9 \\

% 2018 ICRA&SqueezeSeg\cite{wu2018squeezeseg} &30.8 &68.3 &18.1 &5.1 &4.1 &4.8 &16.5 &17.3 &1.2 &84.9 &28.4 &54.7 &4.6 &61.5 &29.2 &59.6 &25.5 &54.7 &11.2 &36.3\\

% 2019 ICRA&SqueezeSegV2\cite{wu2019squeezesegv2} &39.6 &82.7 &21.0 &22.6 &14.5 &15.9 &20.2 &24.3 &2.9 &88.5 &42.4 &65.5 &18.7 &73.8 &41.0 &68.5 &36.9 &58.9 &12.9 &41.0\\

% 2019 IROS&RangNet++\cite{milioto2019rangenet++} &52.2 &91.4 &25.7 &34.4 &25.7 &23.0 &38.3 &38.8 &4.8 &91.8 &65.0 &75.2 &27.8 &87.4 &58.6 &80.5 &55.1 &64.6 &47.9 &55.9\\

% 2020 CVPR&PolarNet\cite{zhang2020polarnet} &54.3 &93.8 &40.3 &30.1 &22.9 &28.5 &43.2 &40.2 &5.6 &90.8 &61.7 &74.4 &21.7 &90.0 &61.3 &84.0 &65.5 &67.8 &51.8 &57.5 \\

% 2021 CVPR&Cylinder3D\cite{zhu2021cylindrical} &68.9 &97.1 &67.6 &63.8 &50.8 &58.5 &73.7 &69.2 &48.0 &92.2 &65.0 &77.0 &32.3 &90.7 &66.5 &85.6 &72.5 &69.8 &62.4 &66.2 \\

% 2021 CVPR&(AF)$^2$-S3Net\cite{cheng20212} &69.7 &94.5 &65.4 & {86.8} &39.2 &41.1 & {80.7} &{80.4} &74.3 &91.3 &68.8 &72.5 &53.5 &87.9 &63.2 &70.2 &{68.5} &53.7 &61.5 & {71.0}  \\

2022 CVPR &PVKD\cite{hou2022point} &71.2 & {97.0} &
 {67.9} &69.3 &53.5 &60.2 &75.1 &73.5 &50.5 &91.8 &70.9 &77.5 &41.0 &92.4 &69.4 & {86.5} &73.8 & {71.9} &64.9 &65.8\\

2022 ECCV &2DPASS\cite{yan20222dpass} &72.9 &97.0 &63.6 &63.4 &61.1 &61.5 &77.9 & {81.3} & {74.1} &89.7 &67.4 &74.7 &40.0 & {93.5} & {72.9} &86.2 & {73.9} &71.0 &65.0 &70.4\\

2022 NIPS & Point Transformer V2\cite{wu2022point} &72.6 &-- &-- &-- &--&--&-- &-- &--&--&--&--&--&--&-- &--&-- &-- &-- &-- \\

% 2022 TMM &   \\

2023 TITS &SAT3D\cite{ibrahim2023sat3d} &{61.3} &{94.5} &{42.1} &45.6 &{21.6} &{39.4} &{63.4} &{61.2} &18.6 &91.8 &68.6 &77.3 &27.2 &91.8 &67.8 &85.8 &70.3 &71.5 &60.3 &64.9  \\

2023 ICCV& RangFormer\cite{kong2023rethinking} & {73.3} &96.7 & {69.4} & {73.7} & {59.9} & {66.2} &78.1 &{75.9} &58.1 & {92.4} & {73.0} & {78.8} & {42.4} &92.3 &70.1 & {86.6} &73.3 & {72.8} & {66.4} &66.6\\

2024 CVPR& Point Transformer V3\cite{wu2023point}$*$ &\underline{74.2} &-- &-- &-- &--&--&-- &-- &--&--&--&--&--&--&-- &--&-- &-- &-- &--\\
\textcolor{black}{2024 TIM} &\textcolor{black}{EPNet\cite{deng2024ep}} &\textcolor{black}{67.4} &-- &-- &-- &--&--&-- &-- &--&--&--&--&--&--&-- &--&-- &-- &-- &--\\
\textcolor{black}{2025 IEEE ITJ} & \textcolor{black}{BeyondPoints\cite{luo2025beyondpoints}} &\textcolor{black}{73.4} &-- &-- &-- &--&--&-- &-- &--&--&--&--&--&--&-- &--&-- &-- &-- &--\\

\textcolor{black}{2025 TGRS} & \textcolor{black}{CSFNet\cite{luo2025csfnet}} &\textcolor{black}{71.5} &\textcolor{black}{97.6} &\textcolor{black}{61.5} &\textcolor{black}{88.5} &\textcolor{black}{96.5}&\textcolor{black}{82.8}&\textcolor{black}{82.4} &\textcolor{black}{93.6} &\textcolor{black}{2.2}&\textcolor{black}{94.8}&\textcolor{black}{52.1}&\textcolor{black}{83.0}&\textcolor{black}{0}&\textcolor{black}{93.4}&\textcolor{black}{73.7} &\textcolor{black}{89.1}&\textcolor{black}{72.8} &\textcolor{black}{75.6} &\textcolor{black}{66.3} &\textcolor{black}{52.3}\\

\textcolor{black}{2025 TCSVT} & \textcolor{black}{NUCNet\cite{wang2025nuc}} &\textcolor{black}{73.6} &\textcolor{black}{97.4} 
&\textcolor{black}{70.9} 
&\textcolor{black}{74.9} 
&\textcolor{black}{60.2}
&\textcolor{black}{63.6}
&\textcolor{black}{79.2} 
&\textcolor{black}{77.6} 
&\textcolor{black}{54.6}
&\textcolor{black}{91.6}
&\textcolor{black}{71.5}
&\textcolor{black}{77.7}
&\textcolor{black}{45.5}
&\textcolor{black}{92.8}
&\textcolor{black}{70.9} 
&\textcolor{black}{86.9}
&\textcolor{black}{75.3} 
&\textcolor{black}{72.6} 
&\textcolor{black}{66.9} 
&\textcolor{black}{68.9}\\

% \textcolor{black}{2025 TITS} & \textcolor{black}{SegNet4D\cite{luo2025semantic}} &\textcolor{black}{71.5} &\textcolor{black}{97.6} &\textcolor{black}{61.5} &\textcolor{black}{88.5} &\textcolor{black}{96.5}&\textcolor{black}{82.8}&\textcolor{black}{82.4} &\textcolor{black}{93.6} &\textcolor{black}{2.2}&\textcolor{black}{94.8}&\textcolor{black}{52.1}&\textcolor{black}{83.0}&\textcolor{black}{0}&\textcolor{black}{93.4}&\textcolor{black}{73.7} &\textcolor{black}{89.1}&\textcolor{black}{72.8} &\textcolor{black}{75.6} &\textcolor{black}{66.3} &\textcolor{black}{52.3}\\

& {SMTransformer}(ours) &  \textbf{74.9} & {97.3} &{66.4} & {65.8}  &  {67.2} &  {68.2} & {80.3} &  {82.7}  &  {76.5} & {92.8} & {71.4} & {82.3} & {38.2} & {92.8} & {70.6} &{86.1} & {74.5} &{70.5} & {67.2} &  {72.3}  \\\hline
\end{tabular}
}\vspace{0mm}
\end{table*}

\textit{Results:} We compare our method with the recent state-of-the-art on S3DIS dataset, using three metrics i.e. mean class-wise intersection over union (mIoU), mean overall accuracy (mAcc), overall accuracy (OA) and parameters (Para.). \textcolor{black}{To ensure a fair comparison with previous state-of-the-art point transformer methods, we reproduced their original code on our own workstation, maintaining the original conditions without using any extra data.}
Results are reported in Table \ref{table: s3dis area5}. Without the test time augmentation trick, our network demonstrates superior performance on  three metrics i.e. 72.6\% mIoU, 78.8\% mAcc and 91.7\% OA, respectively. \textcolor{black}{It outperforms the DeepLA\cite{zeng2025deepla} by 0.7\% mIoU in the term of mIoU.}
\textcolor{black}{With the test time augmentation trick, our network still demonstrates superior performance on three metrics i.e. 73.6\% mIoU, 79.0\% mAcc and 91.8\% OA. Compared to the  Point Transformer V3, the performance of our method exceeds it by 0.5\% mIoU, 0.7\% mAcc and 0.7\% OA, respectively, and our network parameters (i.e. 7.8 million) are reduced by about 68\%.} 
Fig.~\ref{fig:s3disVis}. shows visualizations of our results on S3DIS area 5 in comparison to Point Transformer V2. We can see that our method is more robust to the object boundaries.

The results on ScanNetV2 are illustrated in Table \ref{table: scannet sementic}. Similarly, to ensure a fair comparison, we reproduced the original code of several recent state-of-the-art point transformer methods on our own workstation without using any extra data. These methods include Stratified Transformer\cite{lai2022stratified}, Point Transformer V2 \cite{wu2023point}, ConDAForMer\cite{duan2024condaformer}, Point Transformer V3\cite{wu2023point} and Full Point Transformer \cite{he2023full}. Our network achieved the second-best result with a 76.2\% mIoU, having the fewest parameters (7.8 million) in the compared transformer methods. \textcolor{black}{Compared to the DeepLA, our method exhibits a substantial improvement of +2.1\% mIoU. Although our method performs slightly worse than the Point Transformer V3, our network has only about 17\% of its parameters.} %\textcolor{black}{The results on ScanNetV2 test set are provided in appendix.}

% \vspace{-1mm}
\subsection{Outdoor Semantic Segmentation}

\textit{Datasets:} 
We conduct experiments on two popular datasets:  SWAN \cite{ibrahim2023sat3d} and SemanticKITTI\cite{behley2019semantickitti} dataset. The demanding SWAN dataset comprises 32 sequences of point clouds totalling 10,000 frames and containing approximately 0.9 billion points. Sequences 0 to 23 are allocated for training, while sequences 24 to 31 are designated for testing. The SemanticKITTI provides 22 sequence point clouds consisting of 43,552 frames. Adhering to standard practice, we employ sequences 0 to 10 (excluding 8) for training, use sequence 8 for validation and sequences 11 to 21 for testing. The labels for the test set are exclusively available to the online server, necessitating result submissions for remote evaluation. 
%%%%%%%%%%%%%%%%%%%%%%%%%%%%%%%%%%%%%%%%% TABLE 3 %%%%%%%%%%%%%%%%%%%%%%%%%%%%%%%%%%%%%%%%%%%%%

%%%%%%%%%%%%%%%%%%%%%%%%%%%%%%%%%%%%%%%%% TABLE 4 %%%%%%%%%%%%%%%%%%%%%%%%%%%%%%%%%%%%%%%%%%%%%%

\vspace{0.5mm}
\textit{Network Configurations:} 
For semantic segmentation on SWAN, we opt not to employ voxelization to reduce point resolution; instead, we directly process the raw point cloud data. We set the maximum number of points to 80,000. We use the AdamW optimizer and set weight decay as 0.04. The base learning rate is set as 0.004 and the learning rate is scheduled by the MultiStepLR. We use the same data augmentation as the ones on SemanticKITTI to preprocess the input data. Our model undergoes training and testing phases with a batch size of 16 and 8 distributed across 4 GPUs.  On SemanticKITTI, we set the voxel size as 5cm and the maximum number of voxels to 100,000. We use the AdamW optimizer and weight decay as 0.02. The base learning rate is set as 0.004 and the learning rate is scheduled by the Cosine.  We adopt rotation, flip, scaling, and transformation to augment training data.

\vspace{0.5mm}
\textit{Results:} We present the results for 20 classes of interest in the SWAN test frames in Table \ref{table:swan}. We compare our results to  PointNet++ \cite{wu2022point}, PointConv \cite{wu2019pointconv}, Cylinder3D \cite{zhu2021cylindrical}, PolarNet \cite{zhang2020polarnet}, $\psi$-CNN \cite{lei2019octree}, SAT3D \cite{ibrahim2023sat3d}, \textcolor{black}{Point Transformer \cite{zhao2021point}, Point Transformer V2 \cite{wu2022point
}, and Point Transformer V3 \cite{wu2023point}}. \textcolor{black}{These results are obtained by carefully training the models using the author-provided implements, following the original guidelines for hyper-parameter settings. }As depicted in Table \ref{table:swan}, the mIoU values for these compared methods on the SWAN dataset are lower than those on the semanticKITTI dataset. This discrepancy can be attributed to the heightened complexity of the scenes in the SWAN dataset which was captured in dense central business district are of the city of Perth, Australia. Notably, on this dataset, our method demonstrates the best performance with a remarkable improvement of +4.2\% in mIoU compared to the nearest competitor SAT3D. \textcolor{black}{Besides, compared to the previous point transformers (e.g. Point Transformer V2 and V3), our method exceed them by 2.7\% and 1.3\% mIoU, respectively}. Our method is generally able to show remarkable prediction accuracy for some small objects, including \textit{light poles}, \textit{traffic sign} and \textit{pedestrians}.

% \vspace{-0.5mm}
Table \ref{table: kitti} presents the outcomes of our network alongside results from well-established methods on the SemanticKITTI dataset. Our approach demonstrates commendable performance, achieving 74.9\% mIoU in these compared methods. Notably, compared to the cutting-edge projection-based method RangFormer, which is pre-trained on other datasets, our method exhibits superior performance (+1.6\%) without any pre-training. Furthermore, our proposed method surpasses voxel partitioning and 3D convolution-based techniques, such as 2DPASS, and PVKD. Importantly, \textcolor{black}{compared to the previous state-of-the-art methods, our method outperforms CSFNet and NUCNet by 3.4\% mIoU and 1.3\% mIoU, respectively. Our model showcases a remarkable understanding of certain small object categories, including \textit{poles}, \textit{traffic signs} and \textit{motorcyclists}.} %Fig.~\ref{fig:semantickitti} shows visualizations of our results on SemanticKITTI validation set.

\vspace{-1mm}
\subsection{Ablation Studies}\label{section4_subsection4}
We conduct ablation studies on the S3DIS dataset to demonstrate the effectiveness of SMTransformer, SAUB and Shared Point Position Encoding. 

\label{section4_subsection4_1}
\vspace{0.5mm}
\textit{1)  Effect of Various Components}:  \textcolor{black}{Table \ref{tab:modules} presents the quantitative results of our proposed modules (i.e., SMTB, SAUB, and SPE) in comparison with the Point Transformer (PT) series. Building on PT, Cases I–III progressively integrate SMTB, SAUB, and SPE. The results show that our design not only improves segmentation accuracy but also reduces computational overhead. In particular, Case III (the full model) achieves 73.6\% mIoU with only 7.8M parameters, 2.65G FLOPs, and an inference time of 34ms, surpassing PT (70.6\% mIoU, 7.8M parameters, 2.80G FLOPs, 36ms) in both accuracy and efficiency. Compared with PTV2 and PTV3, our model attains higher accuracy while requiring significantly fewer parameters (7.8M vs. 12.8M/46.19M) and shorter training time (16h vs. 22h/46.5h).} \textcolor{black}{The SPE makes the model lighter because transformer blocks within the same layer share a single position encoding, eliminating redundant parameters. Additionally, by providing a consistent positional representation, the model can focus on feature relationships rather than memorizing layer-specific positional biases. This design contributes to the model’s parameter
efficiency and promotes a more unified geometric understanding.}

\begin{table}[h]
\centering
\arrayrulecolor{black}
\vspace{-2mm}
\caption{\textcolor{black}{Comparison of model efficiency based on the parameters (M), FLOPs (G), inference time (ms) and training time (h). Para.:Parameter. IT: Inference Time. TT: Training Time. }}
% \vspace{-2mm}
\label{tab:modules}
\scriptsize
\renewcommand\arraystretch{1.4}
\setlength{\tabcolsep}{1.6mm}{
\begin{tabular}{c|ccc|ccccc}
\specialrule{1pt}{0pt}{0pt}
 Methods &{SMTB} &{SAUB} &{SPE}   & mIoU  &Para. &\textcolor{black}{FLOPs} &\textcolor{black}{IT} &TT\\\hline
 \textcolor{black}{PTV3} & & &  & \textcolor{black}{73.1} & \textcolor{black}{46.2} &\textcolor{black}{5.82} &\textcolor{black}{94} &\textcolor{black}{46.5} \\
 \textcolor{black}{PTV2} & & &  & \textcolor{black}{71.6} & \textcolor{black}{12.8} &\textcolor{black}{3.64} &\textcolor{black}{42} &\textcolor{black}{22} \\
 % ST & & &  & 91.9 & 18.8 &1.37 &30 &216 \\
 \textcolor{black}{PT} & &  &  & \textcolor{black}{70.6}  &\textcolor{black}{7.8} &\textcolor{black}{2.80} & \textcolor{black}{36} &\textcolor{black}{20} \\\hline
 I &\checkmark  & &  & 72.9  &9.0 &\textcolor{black}{2.96} &\textcolor{black}{38} &22 \\
 II &\checkmark  &\checkmark &  & 73.4  &10.8 &\textcolor{black}{3.24} &\textcolor{black}{43} &24 \\
 III &\checkmark  & \checkmark& \checkmark & {73.6}  &{7.8} & \textcolor{black}{2.65} &\textcolor{black}{34} &16\\\hline
% Point Transformer V1\\
% Point Transformer V3\\
 
\specialrule{1pt}{0pt}{0pt}

\end{tabular}
}
\vspace{-2mm}
\end{table}

% \begin{table}[h]
% \centering
% \arrayrulecolor{black}
% \vspace{-2mm}
% \caption{Effect of various components on semantic segmentation (S3DIS Area-5).
% ` SMTB': Soft Masked Transformer. 
% % `EPE' denotes \textcolor{black}{the proposed} enhanced position encoding. 
% `SAUB': skip attention-based up-sampling block. `SPE': shared point position encoding strategy. `Para.': network parameters. `Time': training time.}
% % \vspace{-2mm}
% \label{tab:modules}
% \scriptsize
% \renewcommand\arraystretch{1.4}
% \setlength{\tabcolsep}{1.6mm}{
% \begin{tabular}{c|ccc|cccc}
% \specialrule{1pt}{0pt}{0pt}
%  Case &{SMTB} &{SAUB} &{SPE}   & mIoU(\%)  &mAcc(\%) &Para.(M) &Time(h)\\\hline\hline
%  I & &  &  & 70.6 &76.5 &7.8 &- \\
%  % II &\checkmark & & &  & 72.0 &77.8 & 8.4 \\
%  II &\checkmark  & &  & 72.9 &78.2 &9.0 &- \\
%  III &\checkmark  &\checkmark &  & 73.4 &78.8 &10.8 &24 \\
%  IV &\checkmark  & \checkmark& \checkmark & {73.6} &{79.0} &{7.8} &16\\
% \specialrule{1pt}{0pt}{0pt}

% % \checkmark & & &62.85 &53.37\\
% % &\checkmark& & 73.00     & 66.57           \\
% % \checkmark &&\checkmark & 67.24     &59.91           \\
% % &\checkmark&\checkmark  & {\bf 75.12}     &      {\bf 68.71}          \\

% \end{tabular}
% }
% \vspace{-2mm}
% \end{table}

\begin{figure*}[t]
\centering
\includegraphics[width=0.9\textwidth]{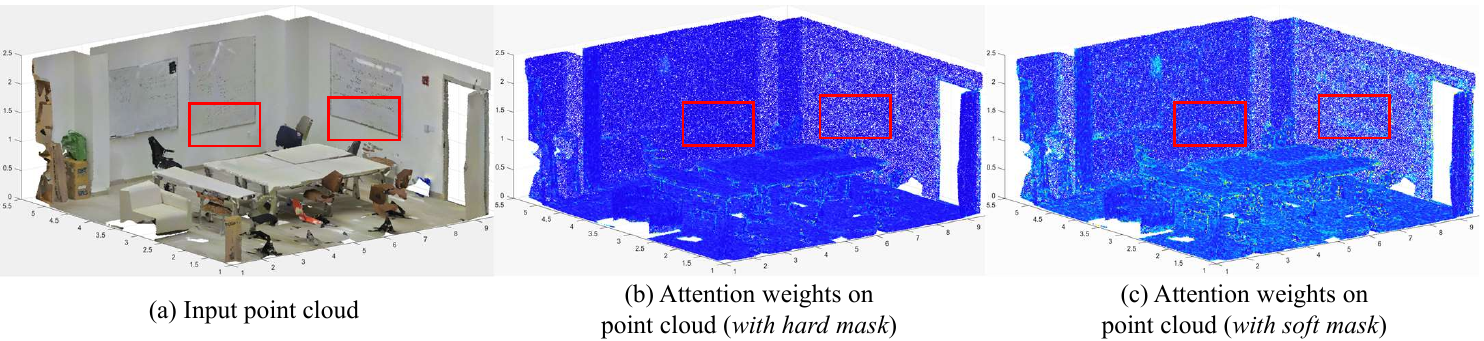}
\vspace{-2mm}\caption{(a) Input point cloud. (b) Attention weights of the hard-masked point transformer. (c) Attention weights of the soft-masked point transformer. The \textcolor{black}{red} boxes emphasize the boundaries of challenging board class where our proposed SMTransformer exhibits superior performance compared to the hard-masked point transformer.} 
\label{fig: softandhard}
\vspace{-4mm}
\end{figure*}

% \vspace{-2mm}
\textit{2) Soft Mask:}
To demonstrate the effectiveness and versatility of the soft mask, we conduct a comparison with the hard mask. The hard mask, represented by binary mask, can be expressed as, 
\vspace{-2mm}
\begin{equation}\nonumber
S(\cdot)=\left\{
\begin{aligned}
&0 \quad K_{ij}^s - Q_i^s <\tau\\
&1 \quad K_{ij}^s - Q_i^s \geq \tau\\
\end{aligned}
\right.,
\end{equation}
where $\tau$ represents the threshold of the mask. When the difference between the task score key $K_{ij}^s$ and the task score query $Q_i^s$ is greater than or equal to the threshold, the corresponding position in the hard mask is set to 1. Otherwise, it is set to 0. Typically, the threshold needs to be optimized for different datasets. Here, we set the $\tau$ as 0.5 on S3DIS through iterative experimentation. \textcolor{black}{Fig. \ref{fig: softandhard} illustrates a comparison of attention weights on the point cloud with soft mask and hard mask, respectively. The point transformer with a mask demonstrates robustness to object boundaries. In particular, the transformer with a hard mask is highly sensitive to certain classes, such as tables and chairs. On the other hand, the transformer with a soft mask not only exhibits common sensitivity to the mentioned classes but also displays high robustness to challenging classes, such as boards.}

\textit{3) Enhanced Position Encoding:} We study the effects of different encoding strategies used for the position encoding in our Transformer encoder. We compare the proposed enhanced position encoding (EPE) with some classical position encoding strategies such as local position encoding (LPE) and global position encoding (GPE). For each of these cases, we test with learnable MLP or nonlearnable Sinusoidal\cite{liu2020closer} position encoding. The results are shown in Table \ref{table: positionencoding}. We can see that the performance of global position encoding is lower than that of local position encoding. The underlying cause is that global position encoding lacks the geometric connection information. When full position encoding strategy is incorporated into the Transformer, the network achieves the highest performance. Our results also show that the MLP encoder is more flexible than Sinusoidal encoder. This indicates our proposed point transformer with full position encoding has more geometric awareness. 
%
% \vspace{-2mm}
\begin{table}[t]
\centering
\vspace{-2mm}
\caption{Segmentation performance of our model on S3DIS area 5 with different position encoding. %LPE: local position encoding, GPE: global position encoding, FGE: proposed full position encoding. MLP: MLP encoder, Sinusoidal: sinusoidal encoder. 
The network does not include SADS block.}
% \vspace{-2mm}
\label{table: positionencoding}
\scriptsize
\renewcommand\arraystretch{1.4}
\setlength{\tabcolsep}{2mm}{
\begin{tabular}{c|c|c||c|c|c}
\specialrule{1pt}{0pt}{0pt}
Encoder &Strategy    &mIoU(\%)  &Encoder &Strategy    &mIoU(\%)    \\ \hline\hline
  Sinusoidal  &LPE  &69.8  &MLP  &LPE & 70.2 \\
  Sinusoidal  &GPE  &68.4  &MLP  &GPE & 69.3 \\
  Sinusoidal  &FPE  &70.5  &MLP  &FPE &\textbf{71.5} \\
\specialrule{1pt}{0pt}{0pt}
\end{tabular}}
\vspace{-4mm}
\end{table}

%%%%%%%%%%%%%%%%%%%%%%%%%%%%%%%%%%%%%%%%%%%%%%%%%%%%%%%%%%%%%%%%%%%%%%%%%%%%%%
%%%%%%%%%%%%%%%%%%%             Sampling Block      %%%%%%%%%%%%%%%%%%%%%%%%%%
\vspace{0.5mm}
\textit{4) Skip Attention-based Up-Sampling Block:} To prove the effectiveness of our proposed skip attention-based up-sampling  block (SAUB), we compare it with two types of up-sampling blocks including the transition up-sampling block (TUB)\cite{zhao2021point} and grid unpooling  block (GUB)\cite{wu2022point}. TUB consists of one projection layer, one interpolation, and one addition operation. GUB consists of one projection layer, \textcolor{black}{one grid  unpooling operation} and one addition operation. The addition operation connects the features from the encoding and decoding layers. Table \ref{table: up-sampling block} presents the performance of our network with various up-sampling blocks. The network employing the SAUB achieves the best performance with mIoU, mAcc, and OA values of 73.6\%, 79.0\%, and 91.8\%, respectively, surpassing the performance of the network using the TUB by a significant margin.

\begin{table}[t]
\centering
\vspace{-2mm}
\caption{Segmentation performance of our model on S3DIS area 5 with different up-sampling blocks.  TUB: transition up-sampling block, GUB: grid unpooling block, SAUB: skip-attention-based up-sampling block. }
% \vspace{-2mm}
\label{table: up-sampling block}
\scriptsize
\renewcommand\arraystretch{1.2}
\setlength{\tabcolsep}{5mm}{
\begin{tabular}{c|c|c|c}
\specialrule{1pt}{0pt}{0pt}
Up-sampling block    &mIoU(\%)  &mAcc(\%) &OA(\%)   \\ \hline\hline
  TUB  &71.0  &76.8  &90.6   \\
  GUB  &72.3  &78.0  &91.2  \\
  {SAUB}(ours)  &\textbf{73.6}  &\textbf{79.0}  &\textbf{91.8}   \\
\specialrule{1pt}{0pt}{0pt}
\end{tabular}}
\vspace{-4mm}
\end{table}

\begin{figure}[h]
\centering
\includegraphics[width=3.in]{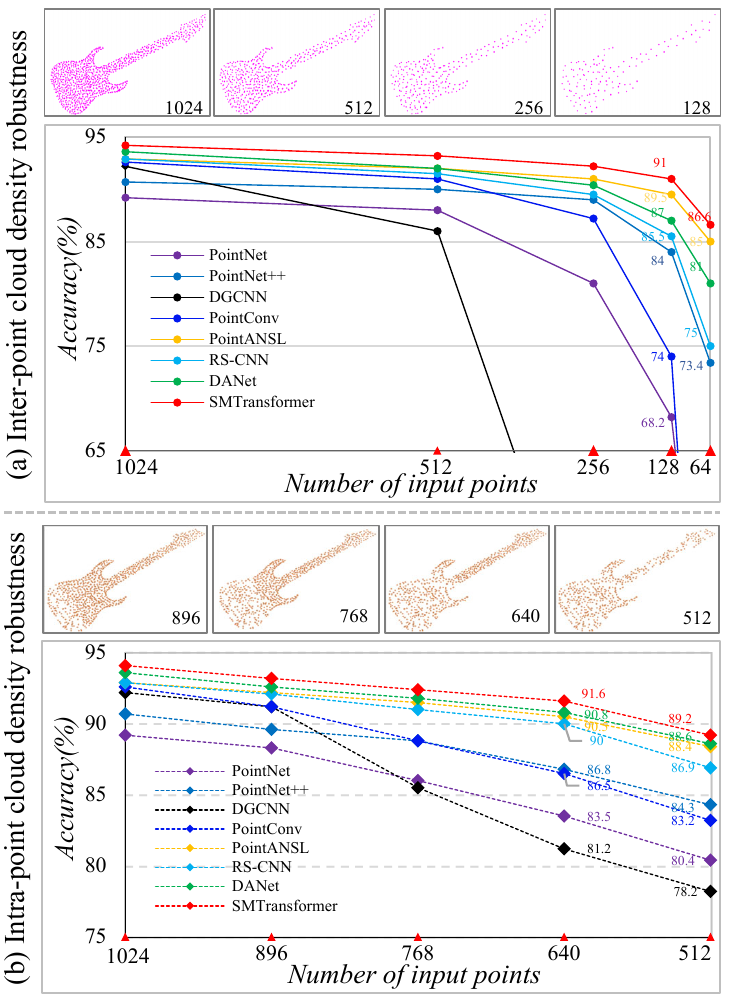}
\vspace{-2mm}\caption{Comparison of classification results on ModelNet40 when points are downsampled to generate (a) inter-point cloud density robustness and (b) intra-point cloud density robustness.} 
\label{robustness_to_density}
\vspace{-2mm}
\end{figure}

\vspace{-4mm}
\subsection{Robustness Analysis}\label{section4_subsection5}
\textit{1) Robustness to Density:} We compare the robustness of our model to inter- and intra-point cloud density with several typical baselines such as \renewcommand{\thefootnote}{\fnsymbol{footnote}} PointNet \cite{qi2017pointnet}, PointNet++ \cite{qi2017pointnet++}, DGCNN\cite{wang2019dynamic}, classical convolutional network such as PointConv \cite{wu2019pointconv}, RS-CNN\cite{liu2019relation}, DANet\cite{he2023danet} and attention network PointASNL\cite{yan2020pointasnl}.  For a fair comparison, all the networks are trained on modelnet40\_normal\_resampled dataset\cite{wu20153d} with 1024 points using only coordinates as the input. To showcase the robustness across inter-point clouds with varying densities, we utilize downsampled points of 512, 256, 128, and 64 as input to the trained model.  To evaluate the robustness of intra-point cloud with various densities, we divide the 1024 points into four equal parts along the $X$ coordinate according to the point number, and then we randomly sample 128 points from each part in sequence. This generates the test samples with 896, 768, 640 and 512 points, respectively. The results are shown in Fig. \ref{robustness_to_density}. Our SMTransformer shows significantly superior robustness, surpassing existing approaches for both inter and intra-point cloud variations. \textcolor{black}{In real-world automation applications, such as robotic perception in warehouses, point density varies drastically due to sensor range, occlusions, or motion. The superior density robustness of SMTransformer suggests that it can maintain reliable perception even under sparse or unevenly distributed point clouds, which is crucial for safety-critical automation tasks.}

\vspace{0.5mm}
\textit{2) Robustness to Transformation:}  To demonstrate the robustness of our SMTransformer, we evaluate its performance on S3DIS and ModelNet40 under a variety of perturbations in the test data, including permutation, translation, scaling and jitter. As shown in Table \ref{table: robustness}, on S3DIS, Point Transformer and Point Vector have a huge performance drop on scaling transformation. Our method exhibits remarkable stability across diverse transformations. Particularly noteworthy is its stable performance even amidst a 0.2 translation along the $X$, $Y$, and $Z$ axes and jitter. All methods are invariant to permutations. In terms of sensitivity to point scaling, SMTransformer performs relatively better when the scaling range is decreased. Our method achieves the best accuracy under all transformations on both segmentation and classification datasets. \textcolor{black}{Robotic and automation systems frequently encounter geometric perturbations, such as sensor misalignment, scaling effects from varying distances, and translation shifts during motion. The stability of SMTransformer under these transformations indicates strong adaptability to such conditions, making it well-suited for robotic navigation and manipulation tasks where viewpoint and scale continuously change.}

\begin{table}[t]
\vspace{-2mm}
\caption{Robustness study for random point permutations, translation of $\pm$~0.2 in $X,Y,Z$ axis, scaling ($\times$0.8,$\times$1.2) and jittering. Note that this ablation study is without test time augmentation.}
\vspace{-2mm}
\label{table: robustness}
\scriptsize
\centering
\renewcommand\arraystretch{1.2}
\setlength{\tabcolsep}{1.2mm}{
\begin{tabular}{l|c|c|c|c|c|c|c}
\specialrule{1pt}{3pt}{0pt}
\multicolumn{1}{c|}{\multirow{2}{*}{{Methods}}} &\multicolumn{1}{c|}{\multirow{2}{*}{{None}}} &\multicolumn{1}{c|}{\multirow{2}{*}{{Perm.}}}  & \multicolumn{2}{c|}{{Translation}} & \multicolumn{2}{c|}{{Scaling}}                     & \multirow{2}{*}{{Jitter}} \\ \cline{4-7}
\multicolumn{1}{c|}{} &\multicolumn{1}{c|}{}   &  \multicolumn{1}{c|}{}    & +~0.2   & -~0.2   & $\times$ 0.8  &$\times$ 1.2 &  \\ 
\hline\hline
\multicolumn{8}{c}{S3DIS Dataset mIoU(\%)} \\ 
\hline

 % PointNet\cite{qi2017pointnet}   &57.75  &59.71 &22.33 &29.85  &56.24  &59.74 &59.04 \\
 % {MinkowskiNet}\cite{choy20194d}   &64.68  &64.56  &64.59  &64.96  &59.60  &61.93 &58.96   \\
% PAConv\cite{xu2021paconv} &65.63 &65.64  &55.81 &57.42 &64.20  &63.94 &65.12  \\
Point Transformer\cite{zhao2021point}   &70.36 &70.45  &70.44 &70.43 &65.73  &66.15 &59.67  \\
Stratified Transformer\cite{lai2022stratified}   &{71.96} &{72.02}  &{71.99} &{71.93} &{70.42}  &{71.21} &{72.02} \\
% Point Vector\cite{deng2023pointvector}   &{72.29} &{72.29}  &{72.29} &{72.29} &69.34 &69.26 &72.16 \\\hline
{SMTransformer}(ours) &\textbf{72.62}  &\textbf{72.62}  &\textbf{72.83} & \textbf{72.96} &\textbf{72.30}  &\textbf{71.94} &\textbf{72.75}   \\
\hline\hline
\multicolumn{8}{c}{ModelNet40 Dataset OA(\%)} \\
\hline
PointNet++\cite{qi2017pointnet++} & 92.1                  & 92.1                      & 90.7           & 90.8               & 91.2      & 91.0    & 91.0     \\
DGCNN\cite{wang2019dynamic}   &92.5    &92.5  &92.3          &92.3                  & 92.1    & 92.3    & 91.5    \\
PointConv\cite{wu2019pointconv} & 91.8                  & 91.8                      & 91.8           & 91.8       & 89.9    & 90.6      & 90.6     \\\hline
{{SMTransformer}}(ours) & \textbf{94.2}     & \textbf{94.2}   & \textbf{94.1}   & \textbf{94.2}       & \textbf{93.5}     &\textbf{93.9}    &\textbf{92.3}\\
\specialrule{1pt}{0pt}{0pt}

\end{tabular}
}\vspace{-2mm}
\end{table}

\begin{table}[t]
\centering
\vspace{-2mm}
\caption{Robustness to background noise on ScanObjectNN. `obj\_bg', `obj\_nobg' stand for objects with and without noise.}
\vspace{-2mm}
\label{table9}
\scriptsize
\renewcommand\arraystretch{1.2}
\setlength{\tabcolsep}{4mm}{
\begin{tabular}{l|c|c|c}
\specialrule{1pt}{0pt}{0pt}
{Method} &{obj\_nobg} &{obj\_bg}    &{OA drop(\%)} \\\hline\hline
3DmFV\cite{ben20183dmfv}      &69.8 &63.0     & 6.8$\downarrow$ \\
PointNet\cite{qi2017pointnet} &74.4&68.2    & 6.2$\downarrow$\\
PointNet++\cite{qi2017pointnet++}   & 80.2 &77.9   & 2.3$\downarrow$ \\
SpiderCNN\cite{xu2018spidercnn} &76.9&73.7  & 3.2$\downarrow$\\
DGCNN\cite{wu2019pointconv}     & 81.5  &78.1  & 3.4$\downarrow$\\
PointCNN\cite{li2018pointcnn}    & 80.8     &78.5  & 2.3$\downarrow$\\ 
{SMTransformer}(ours) &\textbf{88.6} &\textbf{87.3}  & \textbf{1.3$\downarrow$} \\
\specialrule{1pt}{0pt}{0pt}
\end{tabular}
}
\vspace{-4mm}
\end{table}

\textit{3) Robustness to Noise:} To assess  the robustness of SMTransformer to noise, we conducted experiments using the  PB\_T50\_RS variant of ScanObjectNN dataset, measuring the performance with and without background noise (denoted as `obj\_bg' and `obj\_nobg' respectively). Table \ref{table9} presents a comparative analysis between our model and several baselines from \cite{uy2019revisiting}. 
We observe that the overall accuracy of all networks diminishes when trained and tested under conditions involving background noise. However, our model achieves the highest accuracy, exhibiting the smallest performance drop of 1.3\% OA from the `obj\_nobg' variant to the `obj\_bg' variant, surpassing all other networks in comparison. \textcolor{black}{In industrial and robotic environments, raw 3D scans often contain background clutter and measurement noise. The robustness of SMTransformer to noisy data demonstrates its potential for reliable operation in unstructured environments, such as factory floors or outdoor scenes, where clean point clouds cannot be guaranteed.}
% \vspace{-2mm}

% \vspace{-4mm}

\vspace{-2mm}
\section{Conclusion}\label{sec_conclusion}

\begin{comment}
    We proposed a novel soft-masked transformer to explore contextual and task information of point clouds simultaneously. We propose a skip attention-based up-sampling block to fuse the features from distinct resolution points across various encoding layers. We propose a shared position encoding strategy.  Using the above-proposed module, we designed an SMTransformer network that evaluated their performance across various tasks, such as indoor semantic segmentation, outdoor semantic segmentation, and classification. Extensive experiments on challenging benchmarks, as well as thorough ablation studies and theoretical analysis, show the robustness and effectiveness of our method on real-world datasets. As a direction of future research, it will be interesting to observe the effect of adding similar priors or refining the network architecture for point cloud processing tasks.
\end{comment}
In this paper, we introduced a novel Soft Masked Transformer to effectively capture contextual and task-specific information from point clouds. Additionally, we proposed a Skip Attention-based up-sampling block to integrate features from different resolution points across encoding layers. Furthermore, we presented a Shared Point Position Encoding strategy. By incorporating these modules, we constructed an SMTransformer network. Our method was evaluated across various tasks, including indoor and outdoor semantic segmentation. Through extensive experiments on challenging benchmarks, thorough ablation studies and theoretical analysis, we demonstrated the robustness and effectiveness of our approach on real-world datasets. Our contributions significantly advance the state-of-the-art in-point cloud processing. The introduced techniques, including the Soft Masked Transformer, Skip Attention-based Up-sampling block, and Shared Point Position Encoding strategy, provide notable improvements in capturing intricate details and enhancing the performance of point cloud processing tasks. Experimental results on diverse datasets confirmed the efficacy of our proposed approach. As future work, exploring additional priors or refining the network architecture could offer promising avenues to further improve point cloud processing.

\noindent \textcolor{black}{\textbf{Limitations.} Our current framework, like most point-based segmentation methods, operates under a closed-set assumption. When encountering novel classes or unseen geometric shapes, the absence of learned task-level priors may cause the model to misclassify them into the most semantically or geometrically similar known class. Addressing this limitation is crucial for real-world deployment. Future work may focus on integrating class-agnostic feature learning to capture generic geometric patterns, leveraging open-set recognition to detect unfamiliar categories during inference. }

\bibliography{reference}
% \vspace{-4mm}

% \fi
\IEEEpeerreviewmaketitle

\end{document}